*Original Article*

# Enhanced Artificial Neural Networks Using QHAdamW in Air Quality Forecasting

Mary Joy Daniel Viñas

*Computer Studies Department, Technological University of the Philippines, City of Manila, Philippines.*

*Corresponding Author : maryjoy_vinas@tup.edu.ph*

**Abstract -** *The study employed an Artificial Neural Network in combination with the optimized Adaptive Moment Estimation (Adam) algorithm, currently the only AQI forecasting model available in the Philippines. The modified QHAdamW - Quasi-Hyperbolic Momentum (QHAdam) and Adam with decoupled weight decay (AdamW) were both extensions of the Adam optimizer, and both offer unique advantages for training ANN. The proposed QHAdamW optimizer addresses the issues on convergence, generalization, and forecasting performance of Adam. Hyperparameter tuning results revealed that 0.01 and 0.001 were the most effective optimal values for the generalization performance of QHAdamW. The comparative analysis results using seven evaluation metrics revealed that the error value range is lower, and the regression coefficient, having a value approximately equal to 1, improved the model accuracy performance. Likewise, the model converges to a satisfactory level of performance with the convergence performance results of lower loss values as obtained from training and validation losses. Based on data from a real-time air quality tracking station in Manila, a feed-forward neural network is used to predict the AQI of PM2.5 and PM10 separately. This model can be used to forecast Particulate Matter (PM), to help the Department of Environment and Natural Resources - Environmental Monitoring Bureau (DENR-EMB) implement a comprehensive air quality management.*

## 1. Introduction

According to the Air Quality Indices (AQI) developed by the Department of Environment and Natural Resources (DENR) in 2020. Monitoring air qualiyu in crucial for protecting human health, safety, and welfare. The DENR utilizes its Environmental Monitoring Bureau (EMB) to monitor air quality, aiming to protect public health from air pollution risks.

The EMB provides statistics on pollution levels and air quality indexes from both the past and present. Despite the proliferation of research on monitoring air pollution, limited attention has been given to forecast air quality indices. The average annual exposure of Filipino to air pollutants stands at 19 µg/m3, which exceeds the World Health Organization's (WHO) recommended threshold by 1.9 times.

Like weather forecasting, air quality is carried out by local specialist using data and expertise. An air quality forecasting model can be used in this situation. Predicting air quality using available data is known as air quality forecasting. While weather forecasts inform and alert the public about current or upcoming weather conditions, air quality forecasts focus on predicting pollutants and warning about harmful contaminants. To mitigate the consequences of acute air pollution events and warn the public about potential health hazards posed by poor air quality, a predictive approach is crucial [1]. Accurate short-term forecasts of AQI- particularly for PM2.5 (*particulate matter with a diameter less than 2.5 micrometers*) and PM10 (*particulate matter with a diameter less than 10 micrometers*) - are vital for providing early warnings, health advisories of the at-risk populations, and aiding operational planning by local agencies.

However, in the Philippines, existing practices primarily concentrate on observation and reporting, with forecasting systems limited in their convergence (*the point where themodel's prediction stabilizes and stom improving significantly*) and methodological diversity.

While many international studies have shown the promise of machine learning and deep learning in AQI forecasting, there is a gap in applying and evaluating recently proposed optimizer innovations within Artificial Neural Network (ANN) models [2]. ANN structural models can generalize and tolerate errors, making them more adaptable and efficient in problem-solving. ANNs are capable of describing complex nonlinear interactions and generalizing new data.

In the last few years, air quality monitoring and forecasting has become significantly more precise with the advent of various techniques. Advanced forms of traditional techniques, such as Back Propagation Neural Networks (BPNNs) [3], have led to very accurate predictions of daily levels of pollution. Unfortunately, these techniques typically are not very robust because of issues with convergence speed and susceptibility in initial weights. On the other hand, the aggregated Long Short-Term Memory (LSTM) [4] uses sub-neutral network structures that improve its performance, but the long-term average values assigned to the model are not very concentrated due to low accuracy in long-term predictions and when working in resource-constrained environments. The Bayesian Uncertainty Processor [5] and Ensemble Recurrent Neural Networks [6] are very good when they are trained on new unseen data (generalization) and accuracy, and can be improved further with the addition of human activity. However, these techniques are highly reliant on the use of extensive feature engineering and do not perform well when processing very noisy datasets collected in urban environments. Other techniques [7, 8], inspired by nature, such as Particle Swarm Optimization (PSO), have been incorporated in ANN-based systems [9]. Similarly, evolutionary algorithms [10] can be adapted to address problems, but typically local minima before achieving convergence. Alternatively, the use of ANN-based genetic algorithms [11] provides a more reliable methodology that does not have this problem.

The limitations found in convergence stability, susceptibility to local minima, and weight decay of the existing algorithms used to produce air quality forecasts require an efficient optimizer for generating accurate results. Thus, this study develops a new optimize, the Quasi-Hyperbolic Adam with decoupled Weight decay (QHAdamW), which provides a unique contribution to ANN-based forecasting in the Philippines.

In training a neural network, internal parameters such as weights and biases must be adjusted to close the gap between the expected responses from the neural network and the actual output produced [12]. The selection of optimal weights, as well as a learning rate, will determine how well the neural network converges and how well it performs [13]. It should be noted that the accuracy of a neural network will depend on the optimization algorithm used and how much high-quality training data is available for use in training the model.

Gradient descent is an optimization method used to optimize the parameter value of a function by making small incremental adjustments [14] to achieve the least amount of error from the original value (cost function), thereby improving the performance of the overall model by minimizing the error that was created through previous iterations. Adaptive Moment Estimation (Adam) is a common gradient descent technique and has become a popular choice because of its momentum-based updates and flexible learning rate in training neural networks in less time and more efficiently due to its fast convergence, adaptive learning rates and efficient handling of sparse gradients [15, 16]. Adam may not always be the most effective method for controlling obstacles like weight loss, a slow start, and inadequate learning rates that impact generalization and convergence. Some solutions have been implemented, including Quasi-Hyperbolic Adam (QHAdam) [17] and Adam with decoupled weight decay (AdamW) [18], which provides more reliable updates by separating weight decay from gradient updates, resulting in improved regularization.

Although these optimizers have been provided, not many studies have examined the combined use of QHAdam and AdamW, particularly regarding predictions of ambient air quality in urban Philippine settings. The study opens a chance by suggesting QHAdamW as a hybrid optimizer that leverages the strengths of both methods. The research explored this new opportunity by proposing QHAdamW as a hybrid optimizer combining the advantages of both approaches. The central research question is whether QHAdamW outperforms traditional Adam optimizer processes and architectural approaches for achieving convergent, generalized, and accurate forecasting of PM2.5 and PM10 Air Quality Indexes (AQIs). The novelty of this work includes (a) the first application of a combined QHAdam and AdamW optimizer for air quality forecasting in the Philippines; (b) a systematic hyperparameter (adjusting model settings such as learning rate and weight decay to improve accuracy) tuning process and multi-run evaluation that isolates the effects of the optimizer on convergence and generalization; and (c) a practical demonstration using real monitoring data from Environmental Management Bureau National Capital Region (EMB-NCR) for predicting both PM2.5 and PM10 AQI predictions. Unlike prior works that merely modified architectures or applied metaheuristics to weights, this research treats optimizer formulation as the primary experimental variable. It is important to note that the exact combination of the optimized Adam may vary depending on the specific research and experimentation. Different variations or modifications to update rules based on their findings and goals.

Since air pollution problems affect human health and welfare, the research is currently the only ANN -based air quality indicator for forecasting developed in the Philippines. The air pollutants were tracked and anticipated based on the emission inventory from an air quality monitoring station. This study presents the significance of the application of the proposed QHAdamW to the ANN model when providing management strategies for reducing urban pollution. There is no specific research or established updated rules available for the combined QHAdam and AdamW optimization method in the context of air quality forecasting. The combination of these two optimizations is a novel approach, which becomes an optimizer that not only provides independent control over weight decay but also allows the gradient to directly influence the update step. This could potentially offer a different optimization behavior compared to both AdamW and QHadam.

While earlier studies have demonstrated that BPNN, LSTM, Bayesian uncertainty processors, ensemble recurrent networks, and PSO or GA-enhanced ANN model can improve AQI forecasting, each of these approaches also exhibits important limitation and high sensitivity to weight initialization, LSTMs require larger and cleaner datasets and are computationally expensive for short-term AQI prediction, and metaheuristic hybrids are prone to premature convergense and often need extensive tuning to escape local minima.

More so, Adam-based models have been employed as the standard optimizer for deep learning of gradient weight decay, both of which negatively affect generalization to real-world AQi time series. The proposed QHAdamW optimally exploits these deficiencies of Adam/Adam-type optimizers by combining quasi-hyperbolic momentum and decoupled weight decay to stabilize updates to the gradients, accelerate convergence, and improve generalization with respect to noisy air quality estimates from the urban areas of the Philippines.

## 2. Materials and Methods

### *2.1. Data & Procedures*

The research centered on an air quality monitoring station operated by EMB-NCR situated within Mehan Garden, City of Manila [19], making use of state-of-the-art equipment, such as a Teledyne T640 PM analyzer [20]. Since 2020, real-time air quality monitoring has been guaranteed by the Environmental Monitoring Education Division (EMED). The dataset consisted of 14, 328 hourly observations of PM2.5 and PM10 concentrations collected between January 2020 and December 2024 were used in this study.

$$I_p = \frac{I_{Hi} - I_{Lo}}{BP_{Hi} - BP_{Lo}} \left(C_p - BP_{Lo}\right) + I_{Lo} \quad (1)$$

Equation (1) interpolates between the lower and higher breakpoints of the pollutant category to calculate the AQI value corresponding to the concentration of the pollutant. All features were normalized to the range [0, 1] to guarantee consistent training behavior, and linear interpolation was used to handle missing values. Missing values can introduce bias and inaccuracies in the analysis, so it's crucial to handle them appropriately to maintain data integrity.

### *2.2. ANN Architecture*

A subset of Artificial Neural Network (ANN) models the machine learning algorithms stimulated by the architecture of the human brain. ANN models are frequently applied to tasks, including time series forecasting. ANN consists of an input layer, one or more hidden layers, and an output layer. Each layer has multiple neurons. These neurons process input and send information to the next layer, as shown in Figure 1. The forecasting model was designed as a feed-forward multilayer artificial neural network implemented in PyTorch (v2.0). The input features included pollutant concentrations (PM2.5 or PM10) and time indices, while out layer consisted of a single neuron predicting AQI values. The datasets was divided into 70% training, 15% testing subsets. Training was conducted for 100 epochs with a batch size of 32, and early stopping was employed to prevent overfitting by monitoring validation loss. The network architecture comprised two hidden layers with 64 neurons each, and the hyperbolic tangent (tanh) activation function was applied to mitigate vanishing gradient problems and capture nonlinear relationships.

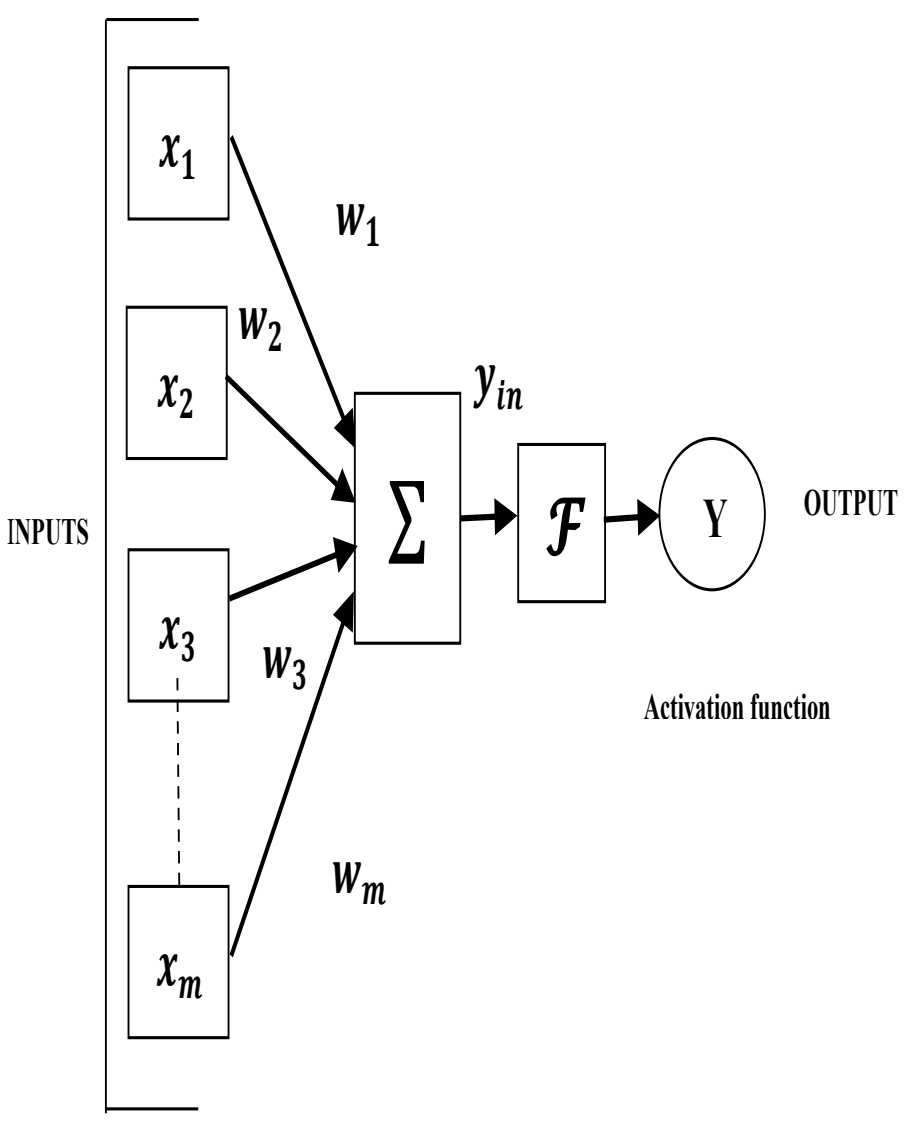

**Fig. 1 The general model of a multilayer artificial neural network**

The model's input is calculated using Equation (2), which generally involves the aggregation of weighted inputs from the previous layer. The input to each neuron in the hidden layer is determined by the input data as well as the weights associated with the connections between neurons in the input layer and the hidden layer. The output of the model is computed using Equation (3), which applies as an activation function to the input. By adding non-linearity, the activation function helps the model identify intricate patterns in the data. The hyperbolic function (tanh) activation function is commonly used in deep neural networks [21]. Tahn effectively addresses the vanishing gradient problem, leading to improved model performance.

$$\mathcal{Y}_{in} = x_1 \cdot w_1 + x_2 \cdot w_2 \dots x_m \cdot w_m = \sum_i^m x_i \cdot w_i \quad (2)$$

$$Y = F(\mathcal{Y}_{in}) \quad (3)$$

### *2.3. Adam Optimizer*

Instead of using SGD, because it can handle sparse gradients and facilitate stochastic Optimization, the Adam optimizer was chosen. Based on the data, the neural network's weights are continuously altered. It is used to find the best weight and save changes. The Adam algorithm is not the same as traditional SGD because it changes all weights at the same learning rate, which stays the same during training. The models are trained using historical data, and their optimal weights are determined using the Adam technique. The formula for enhancing the ANN model is shown in Equation (4).

$$m_t = \beta_1 * m_{t-1} + (1-\beta_1) * g_t v_t = \beta_2 * v_{t-1} + (1-\beta_2) * g_{t2}\, \widehat{m}_t = m_t \;/\; (1-\beta_{1t}) \hat{v}_t = v_t \;/\; (1-\beta_{2t})\, \theta_t = \theta_{t-1} - \alpha * \widehat{m}_t \;/\; \sqrt{(\hat{v}_t)} + \in) \quad (4)$$

The two moving average vectors are: the first moment (mean) of the gradients (shown as m) and the second moment (uncentered variance) of the gradients (v). The algorithm calculates these moving averages for each parameter during training and then uses them to update the parameters. After calculating the bias-corrected values ""m_hat"" and ""v_hat"", the model parameters are updated using the Adam update formula, where ""alpha"" is the learning and the ""epsilon" is a small value to prevent division by zero.

Two optimizers were compared in this study. The baseline Adam optimizer was implemented with default parameters (β1=0.9, β2=0.999, ε=1e-8).

### *2.4. Optimized Adam Algorithms*

AdamW is an enhancement of the Adam optimizer that explicitly addresses the problems with L2 regularization or weight decay that the original Adam optimizer has. Original Adam frequently has trouble controlling weight loss, which might result in less-than-ideal generalization of the design. AdamW addresses this constraint by disconnecting the adaptive learning rates' weight degradation. It is a separation aid in model regularization, lowering the inclination to overfit. AdamW improves performance in generalization, enabling the model to superior to unseen data. In the process of decoupling weight decay, the weights (θ) decay exponentially, as shown in Equation (5).

$$\theta_{t+1} = (1-\lambda)\theta_t - \alpha \nabla f_t(\theta_t) \quad (5)$$

Equation (5) shows how AdamW separates weight decay from gradient updates, which helps prevent overfitting. Equation (6) depicts the updated weights as determined by the Adam algorithm.

$$x_t \longleftarrow x_{t-1} - \alpha \frac{\beta_{1m_{t-1}} + (1-\beta_1)(\nabla f_t + w x_{t-1})}{\sqrt{v_t + \epsilon}} \quad (6)$$

QHAdam is an enhancement of the Adam optimizer that introduces an additional hyperparameter, ""nu."" This modification improves the convergence speed of Adam by incorporating a different update rule for the running average of past gradients. QHAdam improves the optimization process's stability and convergence speed by altering th update rule, especially when noisy and sparse gradients are present. It replaces the moment estimators with quasi-hyperbolic terms. The updates of QHAdam are outlined in Equation (7).

$$\theta_{t+1,i} = \theta_{t,i} - \eta \left[ \frac{(1-v_1).g_t + v_1.\widehat{m}_t}{\sqrt{(1-v_2)g_t^2 + v_2.\hat{v}_t + \in}} \right], \forall t \quad (7)$$

QHAdam combines the advantages of both momentum and simple SGD. It computes a weighted average by adding an instantaneous discount factor (V1) to the current gradient. Next, this number is split into the weighted average of the mean squared gradients and the current squared gradient. The current squared gradient is given more weight by the immediate discount factor V2.

### *2.5. QHAdamW Algorithms*

The main goal of this study is to get better at using the Adam planner to make correct predictions about the AQI. As well as using QHAdam to improve speed and convergence, it also uses AdamW to fix problems with weight loss in the ANN model through regularization. The adaptive learning rate of QHAdam and the better weight decay handling of AdamW work together to make convergence happen faster, generalization better, and Optimization work better.

The proposed QHAdamW optimizer combined quasi-hyperbolic momentum with decoupled weight decay, implemented as a custom optimizer class in PyTorch. Hyperparameter tuning was performed using a grid search across learning rates (0.1, 0.01, 0.001), weight decay values (0.1, 0.01, 0.001), and epsilon constants (0.1,0.01, 0.001). Each configuration was run independently ten times with different random seeds, and results were averaged to assess stability and reproducibility. The researcher describes the following procedures, as seen in Pseudocode 1, by integrating these two approaches and offering revised guidelines for improving the model for air quality forecasting. In this algorithm, AdamW distinguishes between weight decay and loss-based gradient updates within the Adam algorithm. In contrast, QHAdam alters the momentum by increasing the values of beta1 and beta2. These changes cause a delayed buffering process, effectively reducing the variation without unduly decreasing the weight decay.

**Pseudocode 1. QHAdamW Algorithm**

| Step No. | Steps |
|---|---|
| 1 | Initialize the network using random numbers with weights. Read dataset. |
| 2 | Apply the input pattern. |
| 3 | Forecast the output. Split the dataset into a training and a testing dataset. |
| 4 | Let the equations of QHAdamW and AdamW be equal. |
| 5 | $\therefore \theta_t - \gamma(\frac{1}{\sqrt{\hat{v}_t + \epsilon}} \cdot \hat{m}_t + w_t\theta_t) = \theta_t - \gamma[\frac{(1-v_1)g_t + v_1\hat{m}_t}{\sqrt{(1-v_2)g_t^2 + v_2\hat{v}_t + \epsilon}}]$ |
| 6 | Simplify the equation through combining the terms by adding and subtracting coefficients. |
| 7 | Apply the Quotient Property and Square Roots, and cross-multiply the resulting equations. |
| 8 | Apply Transposition Property Equalities to Interchange and simplify more the equation. |
| 9 | Apply the Division property of Equity to get the Theta sub t (Combined QHAdam and AdamW Equation). $\theta_t = \frac{\sqrt{\frac{\hat{v}_t + \epsilon}{(1-v_2)g_t^2 + v_2\hat{v}_t + \epsilon}} \cdot [(1-v_1)g_t + v_1\hat{m}_t] - \hat{m}_t}{w_t}$ |
| 10 | Develop an ANN model, then train and test the model. |
| 11 | Evaluate the model using evaluation metrics. |

### *2.6. Hyperparameter Tuning*

The ideal hyperparameter values are essential for establishing a model's flexibility and complexity after careful analysis of the ANN architectural layers, the number of neurons, the activation function, and the optimizer [24/9/22 9:06:00 AM1]. Table 1 presents the hyperparameters used in the ANN architecture for tuning to achieve optimal performance.

**Table 1. Selected parameters and values to hyperparameter tuning**

| Parameters | Ranges of Values |
|---|---|
| Learning rate | 0.000001(1e-6) to 1.0 |
| Weight Decay | 0.000001(1e-6) to 0.001(1e-3) |
| Epsilon | 0.00000001(1e-8) |

The learning rate determines the size of each step throughout the optimization process [22]. To achieve effective convergence without going above the ideal values, selecting the right learning rate is essential. By punishing large weights and promoting lower weight values for improved generalization, weight decay is a regularization technique that reduces overfitting. Epsilon is a tiny constant that stabilizes the training process and prevents division by zero in optimization algorithms like Adam.

The ANN can be tuned for better performance, convergence and generalization by changing these hyperparameters within the areas they are supposed to be used in. Choosing the correct numerical values for these hyperparameters is essential for making your predictive model both flexible and complex, which will lead to improved AQI predictions for PM2.5 and PM10. The precise selection of the values used to define your hyperparameters will greatly influence how effectively your model will be able to adapt its predictions to new trends in your data set. An appropriately tuned ANN can thus lead to increased accuracy in predicting air quality, and are therefore an important tool in regards to public health and environmental management.

It is not always true that a faster learning rate means faster beginning learning. In practice, if the learning rate is too high, the algorithm may overshoot the ideal values, causing a separation during the training phase. Weight decay is naturally set to a small value, such as 0.001 or 0.01.

Setting it close to 1.0 would impose in a very strong regularization penalty. Weight decay encourages the model to maintain smaller weight values, which helps prevent it from fitting noise in the data and improves its ability to generalize. Epsilon is also included in this context because it is commonly used in Adam algorithms. Epsilon is usually set to a very small constant value.

Table 2 provides the parameter settings used to achieve the optimal performance and convergence with the ANN and proposed QHAdamW, in comparison to the original Adam.

Table 2. Parameters setting for optimization

| Parameters | Value |
|---|---|
| Sequence size | 7 |
| Hidden layers | 2 |
| Hidden nodes | 64, 64 |
| Learning rate | 0.1 |
| Epochs | 100 |
| Weight Decay | 0.0 |
| Activation function | Tanh |
| Number of independent runs* | 10 |

Due to the stochastic nature of the ANN, the model was run independently 10 times. In each run, the results were gathered and evaluated.

### *2.7. Evaluation Metrics*

The forecasting performance and convergence of the ANN model and the suggested QHAdamW in comparison to the original Adam method were assessed using the following assessment metrics. These are used to assess the accuracy of regression models. Equations (8)-(11) show that the model with the lowest MAE, MAPE, MSE, RMSE, and R2 values is the best-fitted model.

Equation (12) shows that the model has a higher correlation [23]. Equations (13) and (14) were included to get a more comprehensive understanding of the model performance.

NRMSE makes more interpretability of the target variable and allows the comparison of model performance across datasets of different sizes, while MAPE, WAPE can be beneficial when you need to give more weight to specific data points or regions of your dataset.

Equation (8) calculates the average of the absolute differences between the actual and anticipated values.

$$MAE\ (y, \hat{y}) = \frac{1}{n_{samples}} \sum_{i=0}^{n_{samples}-1} |y_i - \hat{y}_i| \quad (8)$$

Equation (9) is used to determine the percentage difference between actual and predicted data.

$$MAPE\ (y, \hat{y}) = \frac{1}{n_{samples}} \sum_{i=0}^{n_{samples}-1} \frac{|y_i - \hat{y}_i|}{max\ (\epsilon, |y_i|)} \quad (9)$$

Equation (10) is used to determine the error by comparing the ground truth data with the model's predictions.

$$MSE\ (y, \hat{y}) = \frac{1}{n_{samples}} \sum_{i=0}^{n_{samples}-1} (y_i - \hat{y}_i)^2 \quad (10)$$

Equation (11) was computed using the square root of the average of the squared differences between predicted and actual values on the validation or test datasets.

$$RMSE = \sqrt{\frac{1}{n} \sum_{i=1}^{n} (y_i - \hat{y}_i)^2} \quad (11)$$

To explain the effectiveness of the model's predictions, the variability of the observed data must be measured by Equation (12).

$$R^2(y, \hat{y}) = 1 - \frac{\sum_{i=1}^{n} (y_i - \hat{y}_i)^2}{\sum_{i=1}^{n} (y_i - \bar{y})^2} \quad (12)$$

Where $\bar{y} = \frac{1}{n} \sum_{i=1}^{n} y_i$ and $\sum_{i=1}^{n} (y_i - \hat{y}_i)^2 = \sum_{i=1}^{n} \epsilon_i^2$

NMRSE is commonly used in the evaluation of datasets or forecasting models that exhibit different sizes. The choice of architecture of the ANN, hyperparameters and other influencing factors can significantly affect the model's performance and the resulting NRMSE value, as shown in Equation (13). A lower NRMSE implies a better match between the model and the dataset.

$$NRMSE = \frac{\text{RMSE}}{y_{max} - y_{min}}\ or\ NMRSE = \frac{\text{RMSE}}{\bar{y}} \quad (13)$$

WAPE is particularly useful for data with different scales or levels of importance. This is calculated using Equation (14).

$$WAPE = \frac{\sum_{i=1}^{N} |y_i - p_i|}{\sum_{i=1}^{N} |y_i|} \quad (14)$$

## 3. Results and Discussion

### *3.1. Hyperparameters Tuning Results*

QHAdamW used several hyperparameters that can be adjusted for experimentation and tuning to find the optimal configuration. The study conducted [24] emphasizes the significance of parameter adjustment in reducing the generalization gap when training neural networks.

Through this approach, it is possible to achieve better results that can outperform adaptive gradients algorithms. The parameters were tuned in to determine the most effective values for enhancing the generalization performance of QHAdamW.

The present study focused on experimenting with specific ranges for all parameters, namely 0.1, 0.01, and 0.001, to perform hyperparameter tuning. Learning rate, Weight decay, and Epsilon are the parameters used in this study since these are the most common parameters in various machine learning algorithms.

It is crucial to understand that the aforementioned ranges may exhibit variability due to factors such as the dataset used, the architecture of the model, and the goal of attaining optimal forecasting accuracy.

**Table 3. Hyperparameters optimal values used in forecasting AQI of PM2.5**

| Parameters | | | Test Scores | |
|---|---|---|---|---|
| **learning rate** | **weight decay** | **epsilon** | **RMSE** | **MSE** |
| 0.1 | 0.1 | 0.1 | 7.100597 | 50.4184 |
| 0.1 | 0.1 | 0.01 | 6.358328 | 40.4283 |
| 0.1 | 0.1 | 0.001 | 6.304604 | 39.7480 |
| 0.1 | 0.01 | 0.1 | 6.883581 | 47.3836 |
| 0.1 | 0.01 | 0.01 | 6.269445 | 39.3059 |
| 0.1 | 0.01 | 0.001 | 6.378613 | 40.6867 |
| 0.1 | 0.001 | 0.1 | 6.693702 | 44.8056 |
| 0.1 | 0.001 | 0.01 | 6.330568 | 40.0761 |
| 0.1 | 0.001 | 0.001 | 6.231204 | 38.8198 |
| 0.01 | 0.1 | 0.1 | 7.550536 | 57.0105 |
| 0.01 | 0.1 | 0.01 | 6.230558 | 38.8198 |
| 0.01 | 0.1 | 0.001 | 6.290502 | 39.5704 |
| 0.01 | 0.01 | 0.1 | 7.101930 | 50.4374 |
| 0.01 | 0.01 | 0.01 | 6.228694 | 38.7966 |
| 0.01 | 0.01 | 0.001 | 6.303179 | 39.7300 |
| 0.01 | 0.001 | 0.1 | 7.266685 | 52.8047 |
| 0.01 | 0.001 | 0.01 | 6.277558 | 39.4077 |
| 0.01 | 0.001 | 0.001 | 6.295422 | 39.6323 |
| 0.001 | 0.1 | 0.1 | 6.868913 | 47.1819 |
| 0.001 | 0.1 | 0.01 | 6.311333 | 39.8329 |
| 0.001 | 0.1 | 0.001 | 6.634102 | 44.0113 |
| 0.001 | 0.01 | 0.1 | 7.497867 | 56.2180 |
| 0.001 | 0.01 | 0.01 | 6.311373 | 39.8334 |
| 0.001 | 0.01 | 0.001 | 6.320271 | 39.9458 |
| 0.001 | 0.001 | 0.1 | 7.157912 | 51.2357 |
| 0.001 | 0.001 | 0.01 | 6.256976 | 39.1497 |
| 0.001 | 0.001 | 0.001 | 6.249417 | 39.0552 |

**Table 4. The Optimal values of hyperparameters used in forecasting AQI of PM10**

| Parameters | | | Test Scores | |
|---|---|---|---|---|
| **learning rate** | **weight decay** | **epsilon** | **RMSE** | **MSE** |
| 0.1 | 0.1 | 0.1 | 15.44943 | 238.685 |
| 0.1 | 0.1 | 0.01 | 14.47889 | 209.638 |
| 0.1 | 0.1 | 0.001 | 14.39259 | 207.146 |
| 0.1 | 0.01 | 0.1 | 15.16303 | 229.917 |
| 0.1 | 0.01 | 0.01 | 14.47576 | 209.547 |
| 0.1 | 0.01 | 0.001 | 14.52249 | 210.902 |
| 0.1 | 0.001 | 0.1 | 15.43688 | 238.297 |
| 0.1 | 0.001 | 0.01 | 14.50811 | 210.498 |
| 0.1 | 0.001 | 0.001 | 14.30029 | 204.498 |
| 0.01 | 0.1 | 0.1 | 15.78841 | 249.273 |
| 0.01 | 0.1 | 0.01 | 14.52246 | 210.485 |
| 0.01 | 0.1 | 0.001 | 14.43028 | 208.233 |
| 0.01 | 0.01 | 0.1 | 15.4582 | 238.956 |
| 0.01 | 0.01 | 0.01 | 14.42922 | 208.202 |
| 0.01 | 0.01 | 0.001 | 14.53698 | 211.323 |
| 0.01 | 0.001 | 0.1 | 15.55746 | 242.034 |
| 0.01 | 0.001 | 0.01 | 14.47508 | 209.527 |
| 0.01 | 0.001 | 0.001 | 14.84959 | 220.510 |
| 0.001 | 0.1 | 0.1 | 16.7683 | 281.175 |
| 0.001 | 0.1 | 0.01 | 14.46385 | 209.203 |
| 0.001 | 0.1 | 0.001 | 14.43979 | 208.507 |
| 0.001 | 0.01 | 0.1 | 15.20637 | 231.233 |
| 0.001 | 0.01 | 0.01 | 14.4523 | 208.869 |
| 0.001 | 0.01 | 0.001 | 14.4801 | 209.673 |
| 0.001 | 0.001 | 0.1 | 14.78238 | 218.518 |
| 0.001 | 0.001 | 0.01 | 14.5442 | 211.533 |
| 0.001 | 0.001 | 0.001 | 14.33204 | 205.407 |

Table 3 shows the experimental results for each test run, and the values that yielded the most favorable outcomes with the least errors for PM2.5 were determined to be 0.01 learning rate, weight decay and Epsilon, was found to be the ideal setting. Likewise, Table 4 shows that for PM10, the combination that resulted in the fewest errors included a learning rate of 0.1, along with weight decay and Epsilon both set to 0.001.

### *3.2. Accuracy Performance Results*

During the training phase, the testing set is used to monitor the model's accuracy using the optimal values, the study shows the comparison of the descriptive statistics resuts for both PM10 and PM2. The parameters used in the ANN with the Adam optimizer adhere to the default settings in Keras, while the ANN using QHAdamW uses the optimal values. The network's performance is evaluated on the validation set after each training iteration, known as an epoch. Each model is executed ten (10) times, from which descriptive statistics are computed from these runs. For the study to provide valuable insights into the model's best performance of the AQI from PM10 and PM2.5, these seven evaluation metrics are emphasized: RMSE, MSE, NRMSE, MAE, MAPE, WAPE, and R2, as shown in Table 5. ANN-QHAdamW, based on the comparative analysis results, the error values range in lower values, and the regression coefficient has a value approximately equal to 1, as compared to ANN-Adam. Also, as shown in Table 6, the mean errors of the PM2.5 and PM10 are presented. Looking at all the average improvement results, it is evident that the proposed QHAdamW in the ANN model represents an advantage over the original Adam algorithm. The higher forecasting performance of the proposed ANN-QHAdamW model can be explained by how the optimizer addresses the practical limitations of the state-of-the-art techniques documented in the literature. Traditional ANN and BPNN models, while able to model nonlinear relationships typically converge slowly and are highly sensitive to initial weights, which can lead to suboptimal solutions when trained on noisy AQI data. LSTM-based models improve temporal modelling but require large, clean datasets and higher computational resources; they may be unnecessarily complex for short-term AQI forecasting using medium-length sequences, such as the PM2.5 and PM10 data used in this study.

**Table 5. ANN using QHAdamW and original Adam comparative analysis**

| **Optimizer** | **Particulate Matter 2.5** | | | **Particulate Matter 10** | | |
|---|---|---|---|---|---|---|
| | **QHAdamW** | **Adam** | **Improvement (QHAdamW vs Adam)** | **QHAdamW** | **Adam** | **Improvement (QHAdamW vs Adam)** |
| **RMSE** | 6.22289718 | 6.433495002 | 3.38% | 14.29263 | 14.67882612 | 2.70% |
| **MSE** | 38.72444932 | 41.38985794 | 6.88% | 2042792 | 215.4679362 | 5.48% |
| **NRMSE** | 0.350613302 | 0.362478901 | 3.38% | 0.343548 | 0.827041095 | 140.8% |
| **MAE** | 3.850743054 | 3.930834677 | 2.08% | 9.222333 | 9.370207465 | 1.60% |
| **MAPE** | 0.222978804 | 0.224345811 | 0.61% | 0.215047 | 0.223097462 | 3.74% |
| **WAPE** | 0.210060808 | 0.214429864 | 2.08% | 0.213035 | 0.216450858 | 1.51% |
| **$R^2$** | 0.672339281 | 0.649786353 | 3.35% | 0.583048 | 0.560221103 | 3.92% |

**Table 6. Comparative analysis of average improvement**

| **Optimizer** | **Particulate Matter 2.5** | | | **Particulate Matter 10** | | |
|---|---|---|---|---|---|---|
| | **QHAdamW** | **Adam** | **Improvement (QHAdamW vs Adam)** | **QHAdamW** | **Adam** | **Improvement (QHAdamW vs Adam)** |
| **RMSE** | 6.256375307 | 6.7131995163 | 7.30% | 14.40684 | 14.85216522 | 3.09% |
| **MSE** | 39.142701 | 45.10309024 | 15.28% | 207.5627 | 220.6012272 | 6.28% |
| **NRMSE** | 0.352499542 | 0.37837894 | 7.30% | 0.346294 | 0.836807446 | 141.6% |
| **MAE** | 3.88071525 | 4.044809049 | 4.19% | 9.264983 | 9.546364358 | 3.04% |
| **MAPE** | 0.227709084 | 0.23304825 | 2.34% | 0.221106 | 0.232770375 | 5.28% |
| **WAPE** | 0.211769799 | 0.220647248 | 4.19% | 0.21402 | 0.220520065 | 3.04% |
| **$R^2$** | 0.668800311 | 0.618367433 | 7.54% | 0.576346 | 0.549733532 | 4.62% |

Bayesian uncertainty processors and ensemble recurrent networks enhance generalization but often depend on extensive feature extensive engineering and can struggle when measurement noise and irregular patterns dominate the time series.

There have been positive results from differentiating between metaheuristic-like methodologies that complement ANNs, such as PSO-like global optimizers, but those approaches often have the same types of concerns when combined with ANNs, including premature convergence and getting caught in local minima traps.

Because of the amount of trial-and-error needed when tuning parameters, and having to run a significant number of times in order to see if they have a chance at providing a stable and acceptable result, the computational cost associated with these approaches is a major disadvantage.

The use of aptive methods, such as Adam, are easier to use and have become the baseline for many modern deep learning strategies; however, Adam does not optimally manage the weight decay process and is sensitive to the learning rate, particularly when the initial conditions for the models have been generated using noisy real-world datasets. These qualities could be an impediment to the generalization performance of models trained using Adam, regardless of the degree of decrease in loss during training.

To mitigate these shortcomings of Adam and QHAdamW, two new and complementary algorithms were proposed together with the QHAdamW method. The first component, the quasi-hyperbola, smooths the update to the parameters by taking the instantaneous gradient on each time step and augmenting it with a continuously updated momentum component, which will reduce the rate of fluctuations or oscillations in the model and result in a greater degree of stability during periods of volatility or extreme noise in AQIs.

The second component, AdamW, decouples the weight decay and the update to the parameters based on the instantaneous and averaged gradient used to generate the updated parameters, allowing the model to be improved upon in a more efficient way. Thus, the two components of the QHAdamW optimizers will allow the ANN to converge faster and therefore generalize better compared to using any of the other algorithms mentioned above. The empirical data support the assertion that these algorithms offer the advantages listed above. For every metric of error examined (R2, RMSE, MSE, MAE, etc.) and both PM2.5 and PM10, the QHAdamW model produced consistently better results than the Adam-based ANN did on every type of run that was completed.

Multiple runs and descriptive statistics indicated that these improvements were not random; rather, the gains from each run were consistent across all 10 independent runs.

Additionally, the improved classification accuracy of the AQI categories with QHAdamW was demonstrated to be improved, particularly for PM10; thus, demonstrating that QHAdamW produces not only less numerical error but also produces predictions that can be used for effective decision-making regarding air quality.

Accurate AQI forecasting is crucial for public health; reliable forecasts enable Philippine authorities to take preventive measures and issue health advisories to the general public to protect vulnerable populations, raise awareness and encourage behavior change to collectively contribute to improving air quality.

A comparison of standard deviation values across performance parameters shows that the proposed QHAdamW optimizer is far more stable and consistent than the original Adam optimizer. As shown in Table 7, all evaluation metrics used exhibited much smaller standard deviation with QHAdamW. The computed improvement rates, which span from 142.8% to 847.5%, highlight QHAdamW's remarkable dependability in lowering fluctuations during the training phase.

Regression analysis was performed to determine the correlation between the actual and predicted AQI values using the original Adam and proposed QHAdamW of PM 2.5 and PM10. Likewise, the ANN model for AQI forecasting was perfectly fitted as it obtained the values closer to 1.0. RMSE, MAE, and R2 were considered since these are the widely accepted evaluation criteria for air quality; nevertheless, other popular evaluation criteria are also used - MSE, NRMSE, MAPE and WAPE using QHAdamW obtained the best values on accuracy and forecasting performance.

When compared to the Adam optimizer in the PM0 forecasting model, the QHAdamW optimizer exhibits a lower standard deviation across all performance metrics, suggesting more stable and consistent model behavior. QHAdamW improvements varied from 60.72% to 704.6% with MAE and WAPE showing the most gains. This illustrates QHAdamW's ability for lower prediction errors. For air quality forecasting applications, QHAdamW is a more efficient optimizer than Adam as it improves PM10 prediction accuracy, stability and dependability.

**Table 7. Best performance of PM2.5 and PM10 with QHAdamW as compared to Adam optimizer using standard deviation**

| Optimizer | Particulate Matter 2.5 | | | Particulate Matter 10 | | |
|---|---|---|---|---|---|---|
| | QHAdamW | Adam | Improvement (QHAdamW vs Adam) | QHAdamW | Adam | Improvement (QHAdamW vs Adam) |
| RMSE | 0.200280206 | 0.022828387 | 777.3% | 0.078745 | 0.126558661 | 60.72% |
| MSE | 2.706161796 | 0.285605751 | 847.52% | 2.271056 | 3.770624733 | 66.03% |
| NRMSE | 0.011284278 | 0.001286207 | 777.3% | 0.001893 | 0.007130626 | 276.7% |
| MAE | 0.070919243 | 0.017045943 | 316.0% | 0.023862 | 0.191913933 | 704.3% |
| MAPE | 0.008117346 | 0.00334333 | 142.8% | 0.004469 | 0.012593137 | 181.8% |
| WAPE | 0.003868696 | 0.000929868 | 316.0% | 0.000551 | 0.004433193 | 704.6% |
| R2 | 0.022897754 | 0.002416607 | 847.5% | 0.004635 | 0.007696176 | 66.04% |

### *3.3. Convergence Performance Improvement*

The outcomes of the research will greatly affect the convergence and success of the Fast Forward ANN model that is utilized to determine the AQI for both PM2.5 and PM10. As seen in both Table 8 and Table 9, additional training and validation processes were conducted to confirm that no overfitting occured during comparative analysis of the QHAdamW versus Adam with respect to concentration levels of P2.5 and PM10, as well as the amount of computational time used for each pollutant type. Weight decay aids in in th model's regularization using the suggested QHAdamW optimizer; the gradients' direct influence may result in more effective convergence.

An ANN is trained using optimization methods to minimize a loss function while iteratively updating the model's weights and biases. The QHAdamW optimizer, which uses gradients and moving averages, is used to update the model's weights and biases. Research indicates that QHAdamW shows a greater percentage of improvement compared to Adam.

Additionally, the final validation values obtained using QHAdamW are lower than those achieved with Adam, suggesting that QHAdamW outperforms Adam in model validation. To calculate the percentage error for determining the level of improvement, refer to Equation (16).

$$\text{Improvement (\%)} = \left(\frac{\text{This Study} - (\text{ANN} - \text{Adam})}{\text{This Study}}\right) * 100 \quad (16)$$

The findings of this study have a significantly influence the convergence and performance of the ANN model used in forecasting the PM2.5 and PM10 AQI. In the comparative analysis result between the QHAdamW and Adam algorithms, we monitored the computational time, as well as the training and validation loss, to ensure the model did not overfit, as shown in Tables 10 and 11.

Figures 2 and 3 illustrate the selected concentration pollutants and their forecasted AQI values in an ANN model using the proposed QHAdamW.

**Table 8. Validation loss and the percentage of improvement (PM2.5)**

| Trial | Adam | | | QHAdamW | | |
|---|---|---|---|---|---|---|
| | Initial Value | Final Value | Improvement (%) | Initial Value | Final Value | Improvement (%) |
| 1 | 0.001927 | 0.001792 | 7.00 | 0.002253176 | 0.001751586 | 22.26 |
| 2 | 0.002179 | 0.001788 | 17.95 | 0.002789846 | 0.00173681 | 37.75 |
| 3 | 0.001806 | 0.001806 | - | 0.002253363 | 0.001743632 | 22.62 |
| 4 | 0.001926 | 0.001823 | 5.35 | 0.002839071 | 0.001792299 | 36.87 |
| 5 | 0.001852 | 0.001827 | 1.35 | 0.002671597 | 0.00178512 | 33.18 |
| 6 | 0.001854 | 0.001805 | 2.68 | 0.002456205 | 0.001737607 | 29.26 |
| 7 | 0.001812 | 0.001775 | 2.05 | 0.002231965 | 0.001751316 | 21.53 |
| 8 | 0.003219 | 0.001815 | 43.61 | 0.003639169 | 0.001762703 | 51.56 |
| 9 | 0.001887 | 0.001794 | 4.96 | 0.002795357 | 0.001774075 | 36.53 |
| 10 | 0.001795 | 0.001795 | - | 0.002170886 | 0.001786717 | 17.70 |
| MIN | 0.001795 | 0.001775 | - | 0.002253176 | 0.00175158 | 17.70 |
| MAX | 0.003219 | 0.001827 | 43.61 | 0.002789846 | 0.00173681 | 51.56 |
| AVG | 0.002026 | 0.001802 | 8.49 | 0.002253363 | 0.00174363 | 30.93 |

**Table 9. Validation loss and the percentage of improvement (PM10)**

| Trial | Adam | | | QHAdamW | | |
|---|---|---|---|---|---|---|
| | Initial Value | Final Value | Improvement (%) | Initial Value | Final Value | Improvement (%) |
| 1 | 0.002473 | 0.002185 | 11.65 | 0.003681 | 0.002184 | 40.68 |
| 2 | 0.002232 | 0.002184 | 2.15 | 0.002880 | 0.002183 | 24.21 |
| 3 | 0.002287 | 0.00217 | 5.12 | 0.002647 | 0.002200 | 16.88 |
| 4 | 0.002854 | 0.002202 | 22.85 | 0.002566 | 0.002178 | 15.14 |
| 5 | 0.002373 | 0.002173 | 8.45 | 0.002662 | 0.002204 | 17.22 |
| 6 | 0.002633 | 0.002196 | 16.58 | 0.002387 | 0.002203 | 7.70 |
| 7 | 0.002444 | 0.002189 | 10.42 | 0.002580 | 0.002172 | 15.83 |
| 8 | 0.002318 | 0.002177 | 6.10 | 0.002743 | 0.002182 | 20.44 |
| 9 | 0.002334 | 0.00219 | 6.16 | 0.002767 | 0.002207 | 20.23 |
| 10 | 0.002574 | 0.002196 | 14.70 | 0.003012 | 0.002182 | 27.53 |
| MIN | 0.002318 | 0.002177 | 6.10 | 0.02387 | 0.002172 | 7.70 |
| MAX | 0.002633 | 0.002196 | 16.58 | 0.003012 | 0.002207 | 27.53 |
| AVG | 0.002461 | 0.00219 | 10.79 | 0.002697 | 0.002189 | 18.35 |

**Table 10. QHAdamW and Adam computational time and loss (PM2.5)**

| Parameters | Computational time | Training Loss | Validation Loss |
|---|---|---|---|
| **Adam** | 0.0023119 | 0.001514 | 0.0018825 |
| **QHAdamW** | 0.0021025 | 0.001641 | 0.0017613 |

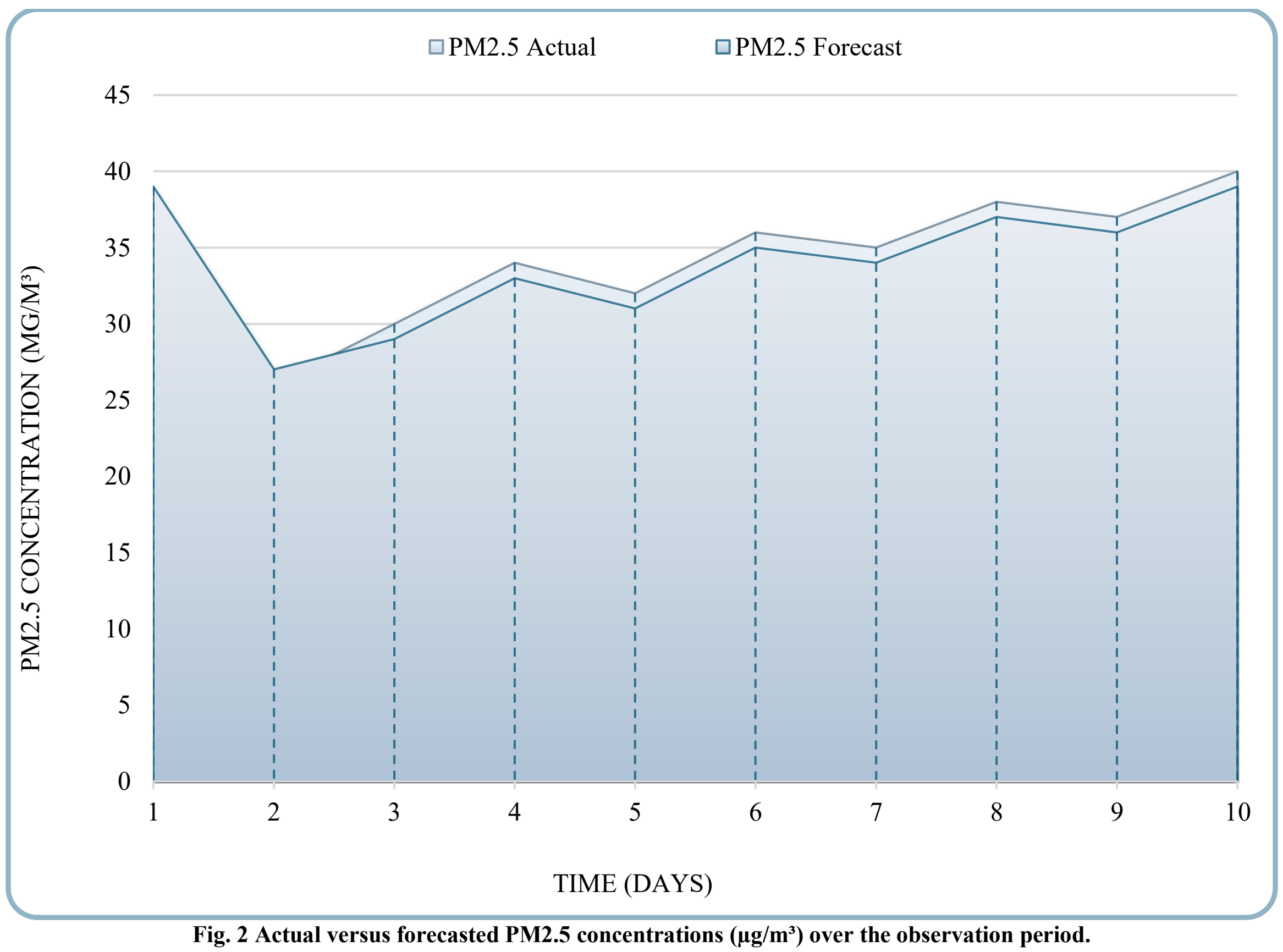

**Fig. 2 Actual versus forecasted PM2.5 concentrations (µg/m³) over the observation period.**

**Table 11. QHAdamW and Adam computational time and loss (PM10)**

| Parameters | Computational time | Training Loss | Validation Loss |
|---|---|---|---|
| **Adam** | 0.002760 | 0.001985 | 0.002177 |
| **QHAdamW** | 0.002161 | 0.002147 | 0.002203 |

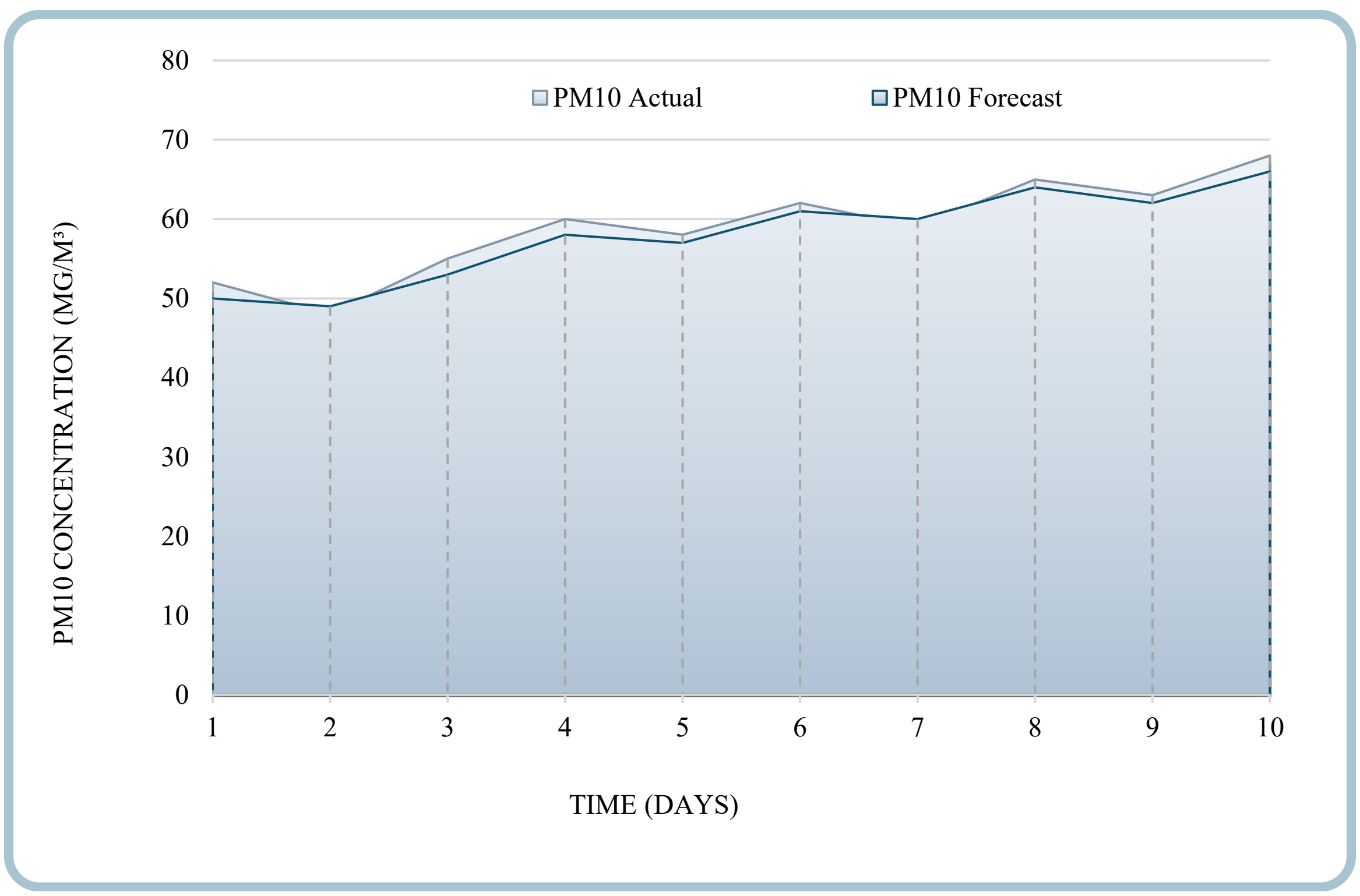

**Fig. 3 Actual versus forecasted PM10 concentrations (μg/m³) over the observation period.**

Figures 4 to 7 illustrate a comparison of training and validation results for the Adam and QHAdamW optimizers over 100 epochs. The primary focus of this investigation is the performance of these optimizers concerning PM2.5 and PM10 AQI values. Based on the training and validation loss metrics observed across 100 epochs, the evaluation shows the benefits and drawbacks of each optimizer in terms of convergence and generalization. An evaluation of how much loss is reduced through both training/validations will help confirm whether a given optimizer produces an appropriate model convergence and optimal performance level from the underlying model parameters. To this end, this investigation considered the relative performances between each option using PM2.5 and PM10 forecasting. After 100 epochs of training and evaluating both the optimizer vs. the QHAdamW optimizer, may provide improvements in loss reduction relative to the Adam model.

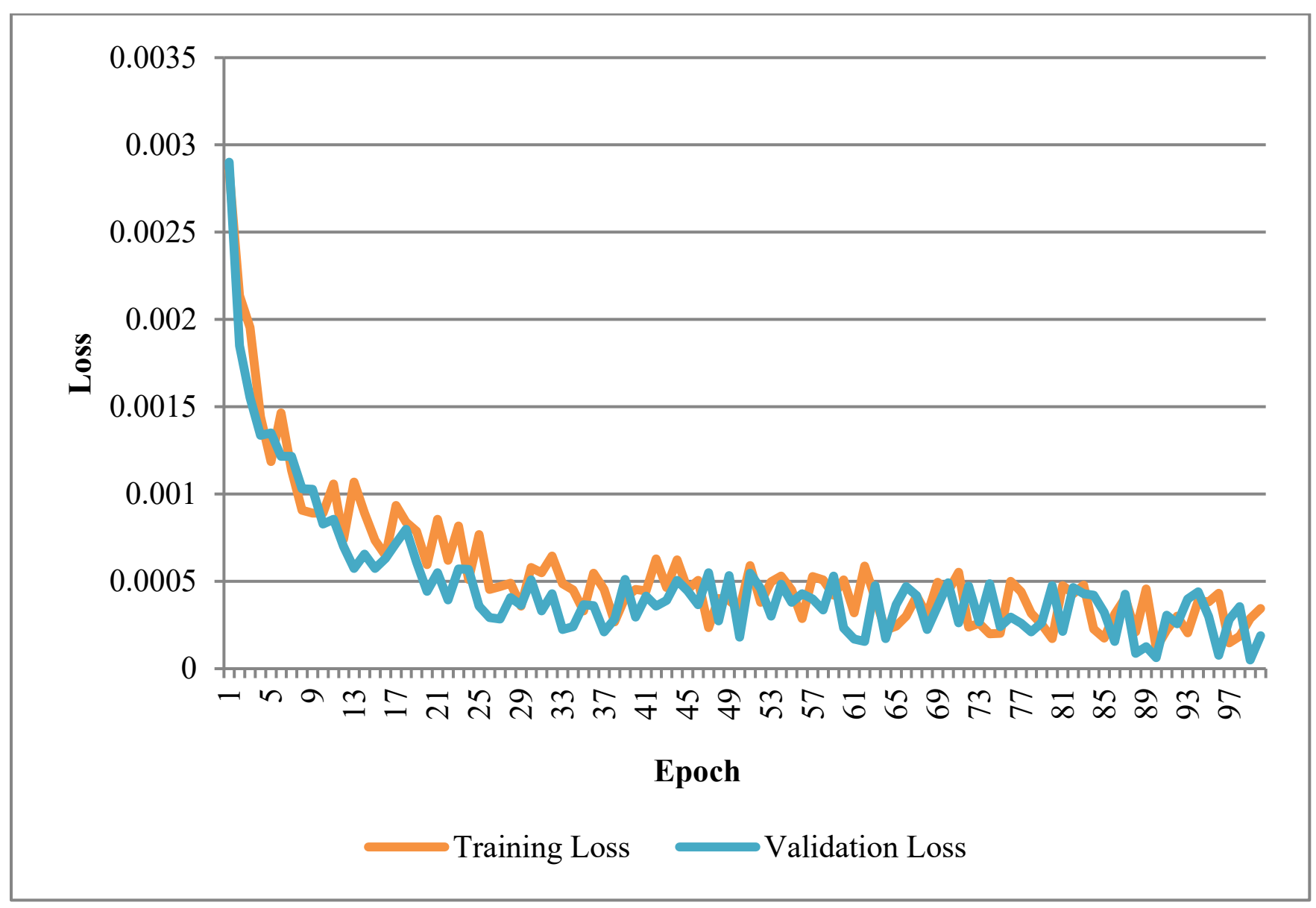

**Fig. 4 Loss with PM2.5 performance and Adam optimizer**

Table 12. Adam training and validation percentage improvement (PM2.5)

| Loss | Initial | Final | %Improvement |
|---|---|---|---|
| **Training** | 0.002664 | 0.001517 | 43.04% |
| **Validation** | 0.001788 | 0.001787 | 0.034% |

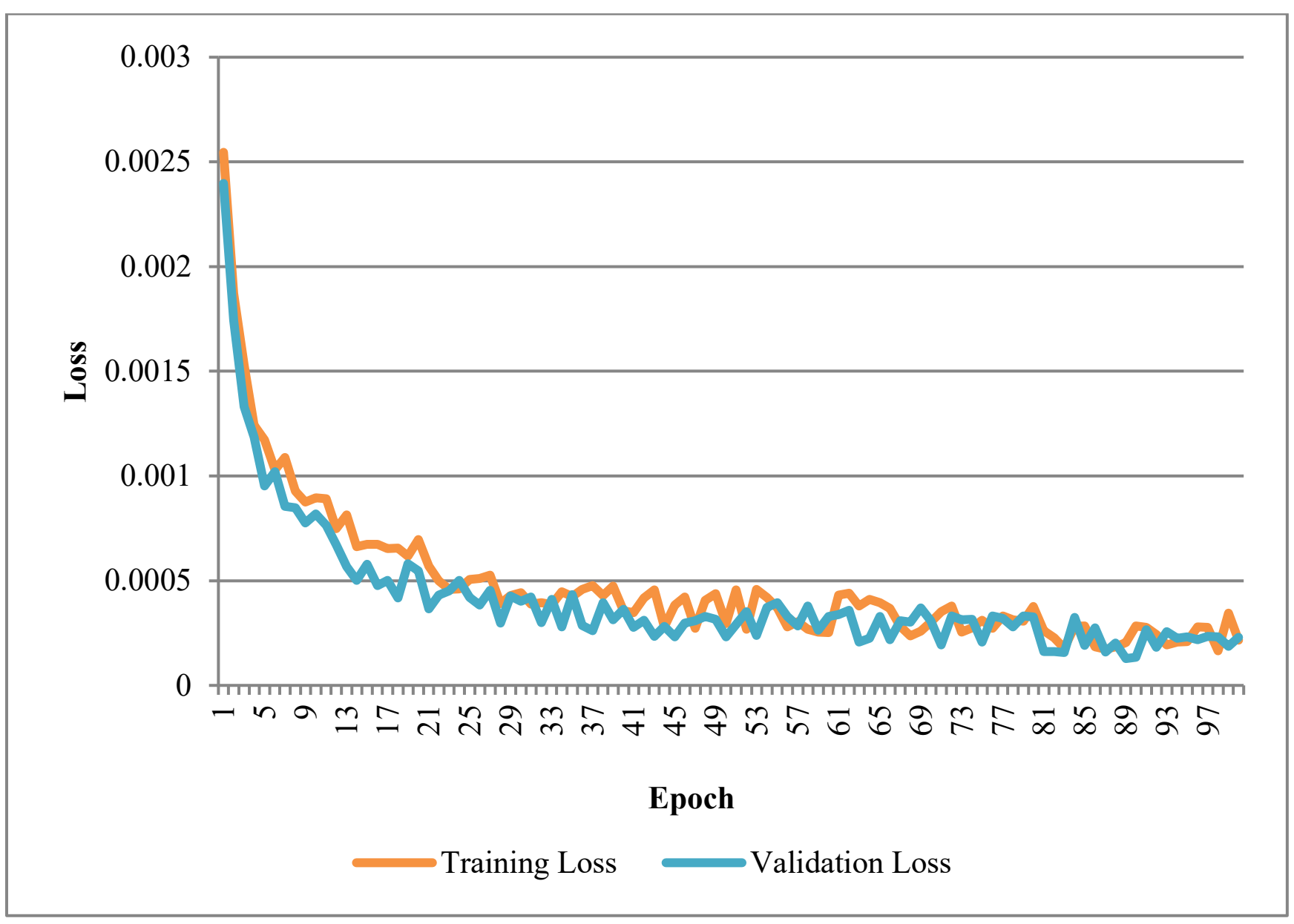

Fig. 5 Loss with PM2.5 performance and QHAdamW optimizer

Table 13. QHAdamW Training and Validation Percentage Improvement (PM2.5)

| Loss | Initial | Final | %Improvement |
|---|---|---|---|
| **Training** | 0.003654 | 0.001977 | 45.88% |
| **Validation** | 0.003227 | 0.001768 | 45.21% |

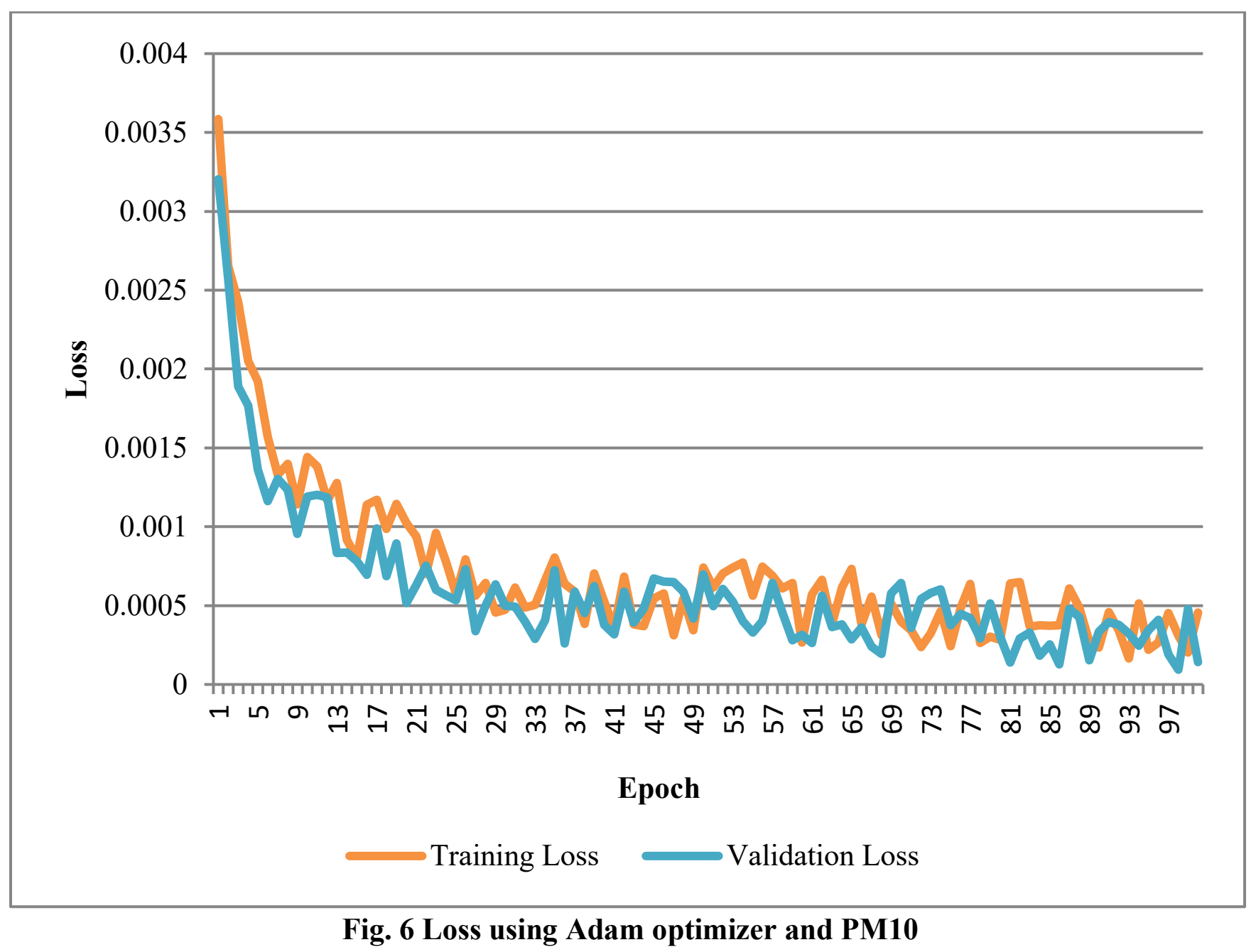

Fig. 6 Loss using Adam optimizer and PM10

Table 14. Adam training and validation percentage improvement (PM10)

| Loss | Initial | Final | %Improvement |
|---|---|---|---|
| Training | 0.003116 | 0.002002 | 35.76% |
| Validation | 0.002223 | 0.002219 | 0.263% |

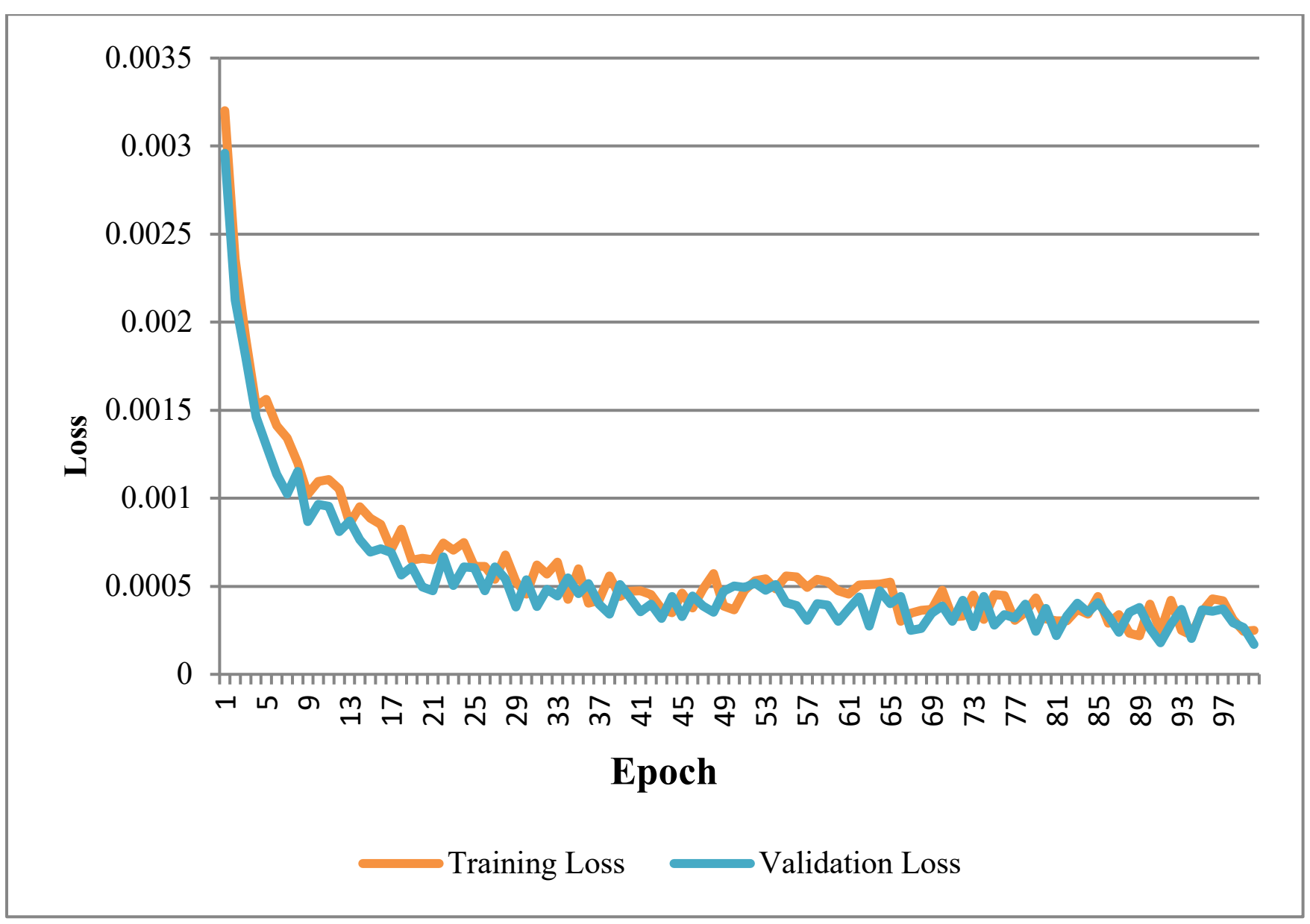

Fig. 7 Loss using QHAdamW optimizer and PM10

Table 15. QHAdamW training and validation percentage improvement (PM10)

| Loss | Initial | Final | %Improvement |
|---|---|---|---|
| Training | 0.002780 | 0.002388 | 14.11% |
| Validation | 0.000283 | 0.002184 | 22.98% |

During the training phase, the Adam optimizer produced a significant decrease in training loss of 43.04%, which indicates that successful learning occurred through the progression of this learning process, whereas validation loss reflected only an extremely small improvement at 0.034 percent, thus suggesting limited abiliy to generalize to new data and indicating some likelihood of overfitted results.

Contrary to Adam, the QHAdamW optimizer produced a greater training loss reduction of (45.88%) compared to the training loss reductions produced by the Adam optimizer. In addition, the QHAdamW produced a significant validation loss improvement of (45.21%), which indicates both (1) that QHAdamW was efficient at learning during the training process, and (2) that QHAdamW has excellent generalization capabilities, thereby (by inference) would likely reduce model overfitting and provide additional strength to model robustness.

Although QHAdamW presented a lower training loss increase of 14.11%, the gotten validation loss difference between the two algorithms was greater than Adam's performance difference at 22.98%. This may seem counterintuitive if only managing to see a 35.76% reduction in PM10 forecasting losses; however, the validity loss difference was only 0.263, with Adam displaying difficulty with generalization error. Therefore, QHAdamW provides better balance (learning /generalization) for PM10 forecasting than Adam. In addition, between the two, QHAdam consistently provides less total validity lossess for both PM2.5 and PM10, indicating an overall greater understanding of the generalization process.

The present study compares the relative abilities of Adam versus QHAdamW to forecast the AQI for the only pollutants measured with AQI data (PM2.5 and PM10). The data demonstrate that, while Adam exhibits strong convergence during training, its validation performance is severely deficient, implying probable overfitting.

On the other hand, QHAdamW not only improves training loss but also significantly decreases validation loss, indicating increased generalization and robustness. QHAdamW's balanced optimization technique improves its reliability for AQI forecasting in complicated urban environments with data noise and unpredictability. The enhanced ANN-QHAdamW forecasting model provides actionable insights for air quality management. By improving the prediction accuracy of PM2.5 and PM10, the model can support DENR-EMB in issuing timely advisories, guide local governments in traffic and industrial regulation, and empower communities – especially vulnerable groups - to make informed decisions about outdoor exposure. The framework is adaptable to other urban centers, offering a scalable tool for sustainable pollution mitigation.

### *3.4. ANN AQI Forecasting Accuracy*

Public health depends on accurate AQI forecasting; honest forecasts allow Philippine authorities to protect vulnerable populations, increase public awareness, and promote behavior change in order to improve air quality. The accuracy applied to machine learning measures the performance of a model by quantifying the number of correct predictions out of the total predictions made, as expressed in Equation (17).

$$Accuracy = \frac{Number\ of\ Correct\ Predictions}{Total\ Number\ of\ Predictions} \quad (17)$$

This metric evaluates how accurately the two models, Adam and QHAdamW, predicted the AQI levels of PM2.5 and PM10. Based on certain ranges, the AQI scores are divided into six groups – good, fair, unhealthy, very healthy, acutely healthy, and emergency, in order to assess how accurate their forecasts are. The highest AQI value indicates the worst air quality conditions, and these categories represent different levels of air pollution to help educate the public. If the category assigned by the models matches the category given to the actual AQI data, it is counted as a correct prediction. In other words, a model is deemed successful if it accurately categorizes the air quality. In this study, the total number of predictions for both PM2.5 and PM10 was 13, 128. Table 9 below displays the number of correct predictions made by the two optimizers.

**Table 16. Number of correct predictions**

| Particulate | Optimizer | Correct Predictions | Accuracy |
|---|---|---|---|
| PM 2.5 | Adam | 6,788 | 51.71% |
| | QHAdamW | 6,965 | 53.05% |
| PM 10 | Adam | 10,403 | 79.24% |
| | QHAdamW | 11,557 | 88.03% |

The results, as shown in Table 16, illustrate that the QHAdamW optimizer outperformed the Adam optimizer. For PM2.5, accuracy increased by 2.53%, from 51.71% with Adam to 53.05% with QHAdamW. The accuracy of PM10 increased by 9.99%, from 79.24% using Adam to 88.03% with QHAdamW. By employing an optimizer that is especially suited to noisy, real-time AQI data, the suggested technique enhances these accuracy increases, which are in line with the behavior shown in previous state-of-the-art research utilizing BPNN, LSTM, and hybrid metaheuristic-ANN models., QHAdamW enhances Optimization for ANN frameworks by making it easier to find a working combination of weights in the model and allows for greater accuracy in the final result, while also being more robust.

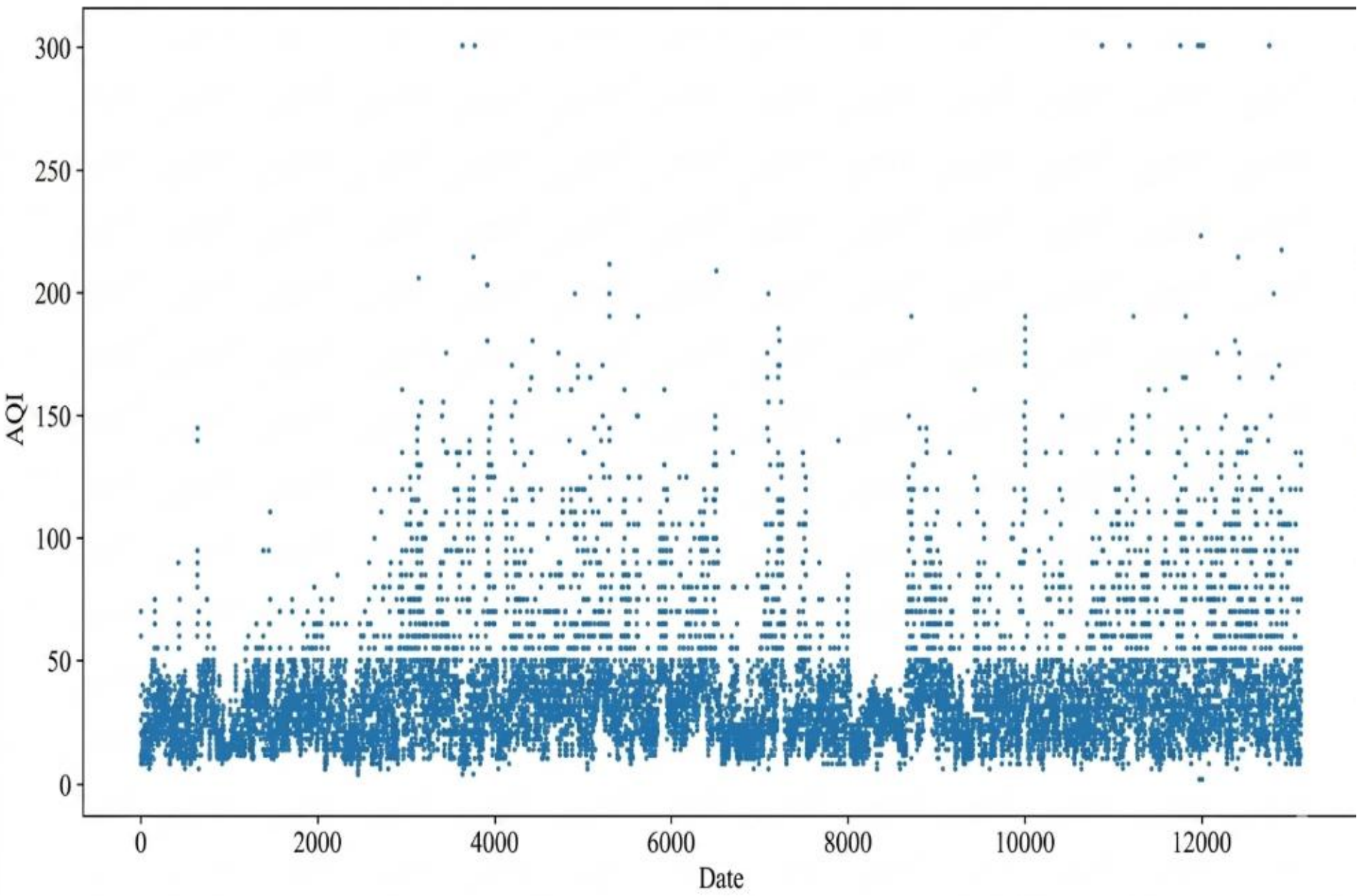

**Fig. 8 ANN-Adam AQI forecasting (PM2.5 AQI)**

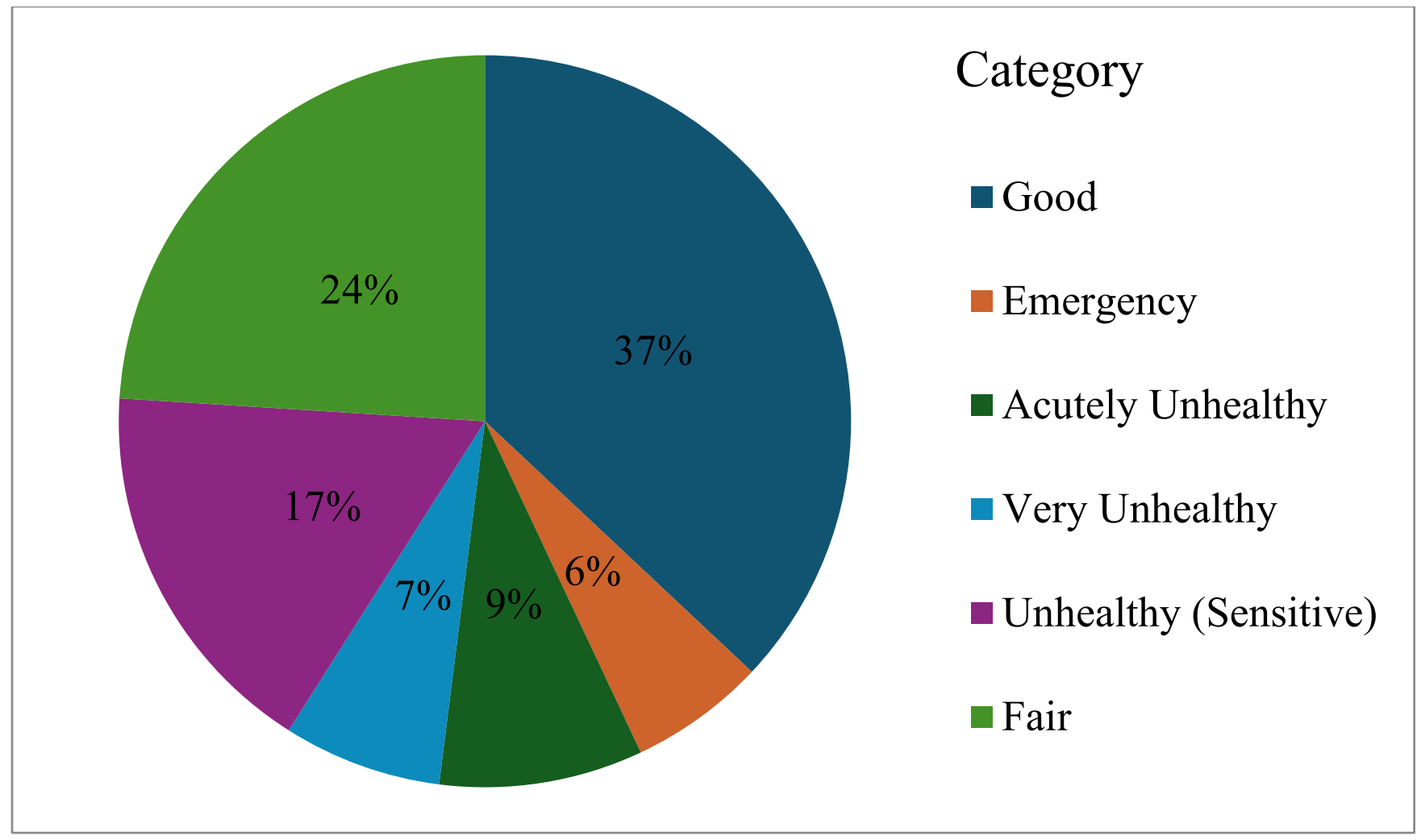

**Fig. 9 Adam prediction per category (PM2.5 AQI)**

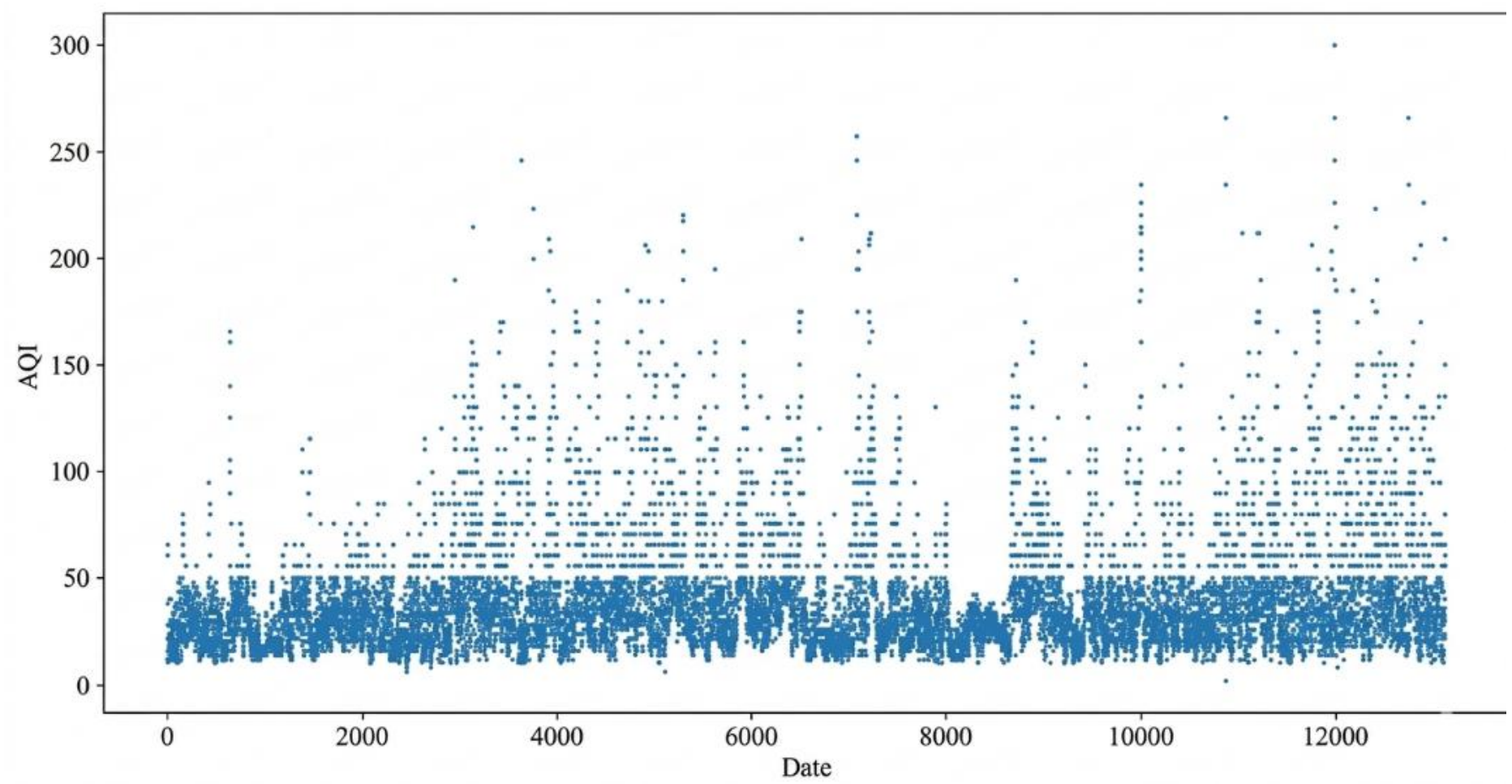

**Fig. 10 ANN-QHAdamW AQI forecasting (PM2.5 AQI)**

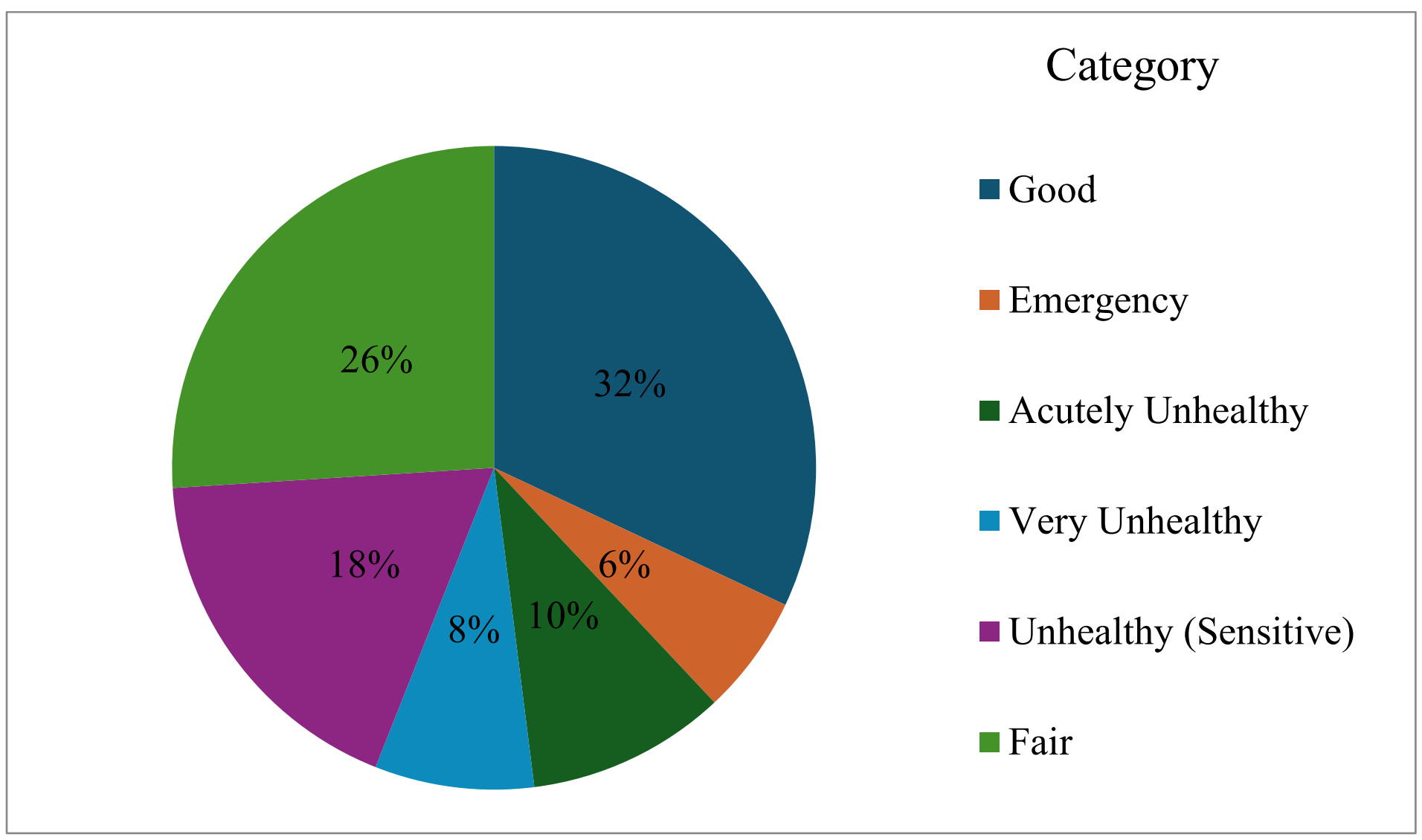

**Fig. 11 QHAdamW prediction per category (PM2.5AQI)**

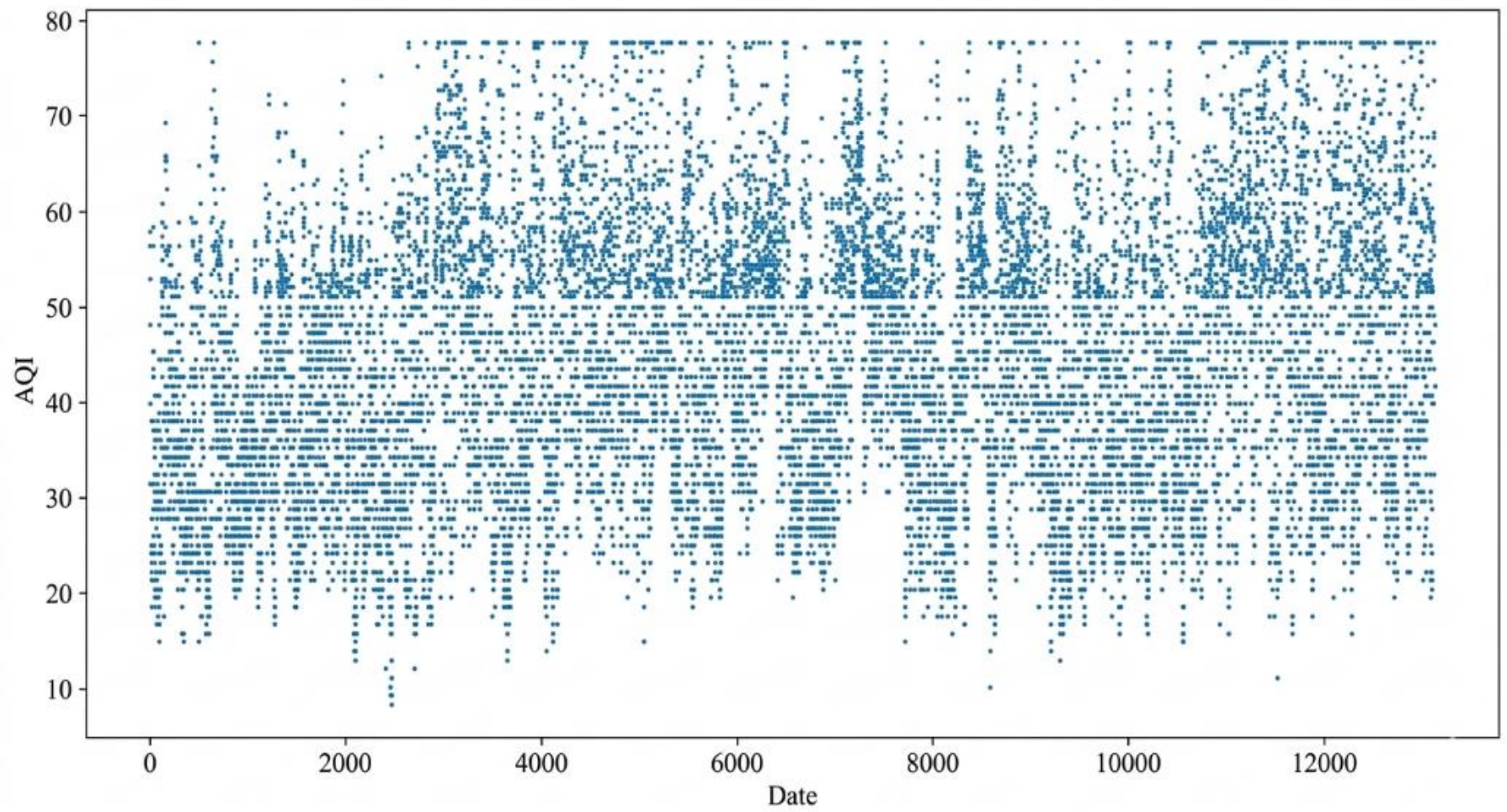

**Fig. 12 ANN-Adam AQI forecasting (PM10 AQI)**

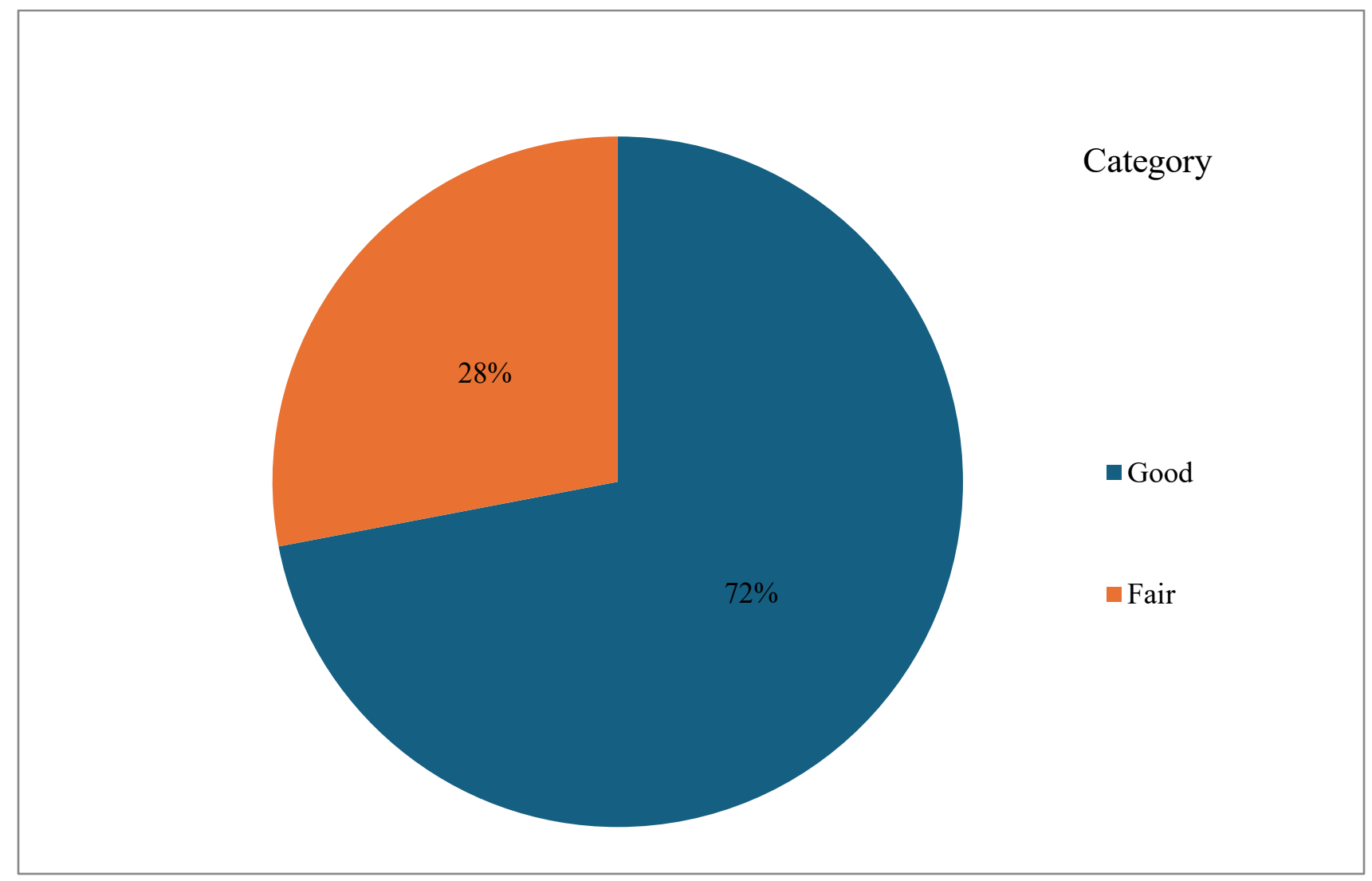

**Fig. 13 Adam prediction per category (PM10 AQI)**

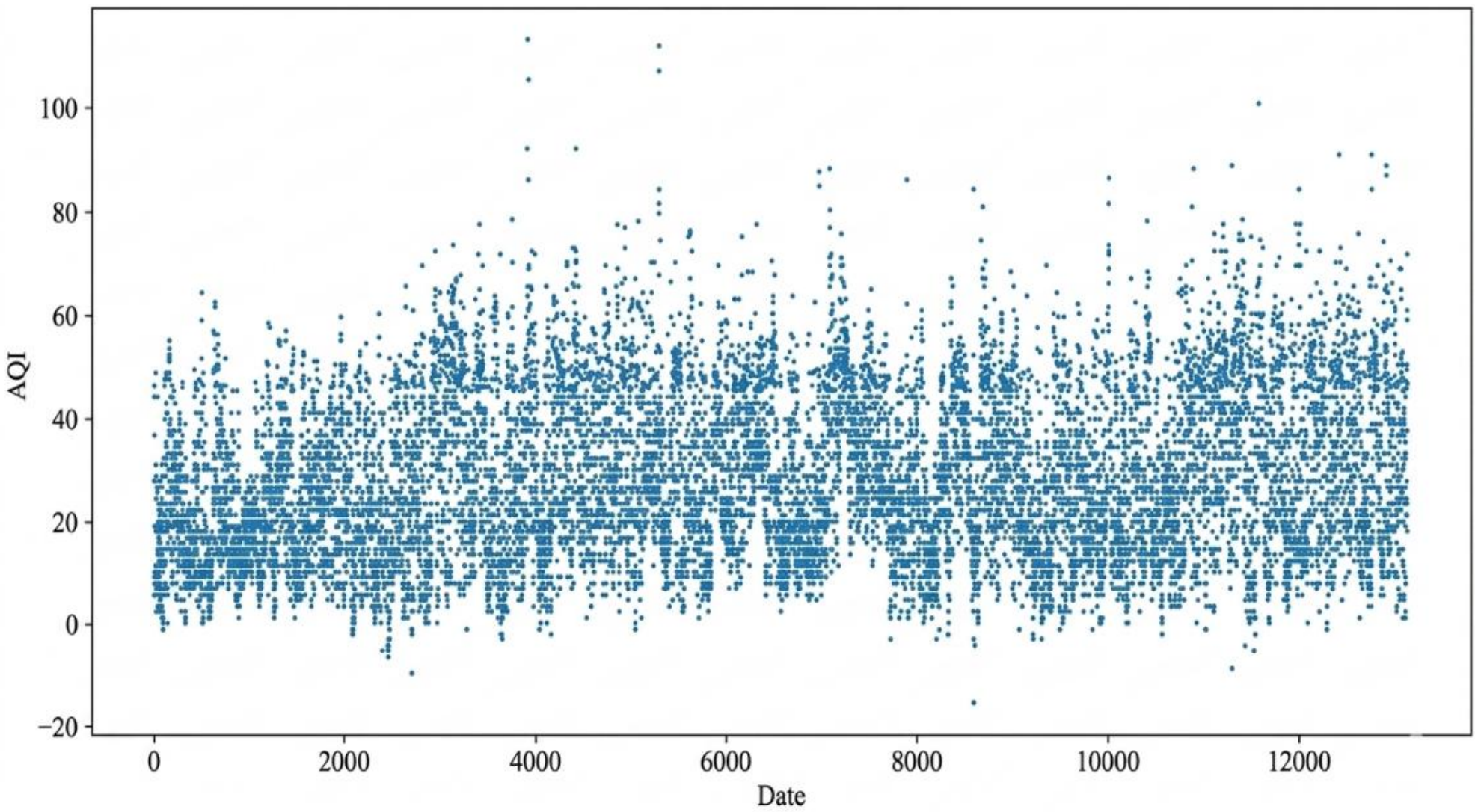

**Fig. 14 ANN-QHAdamW AQI forecasting (PM10 AQI)**

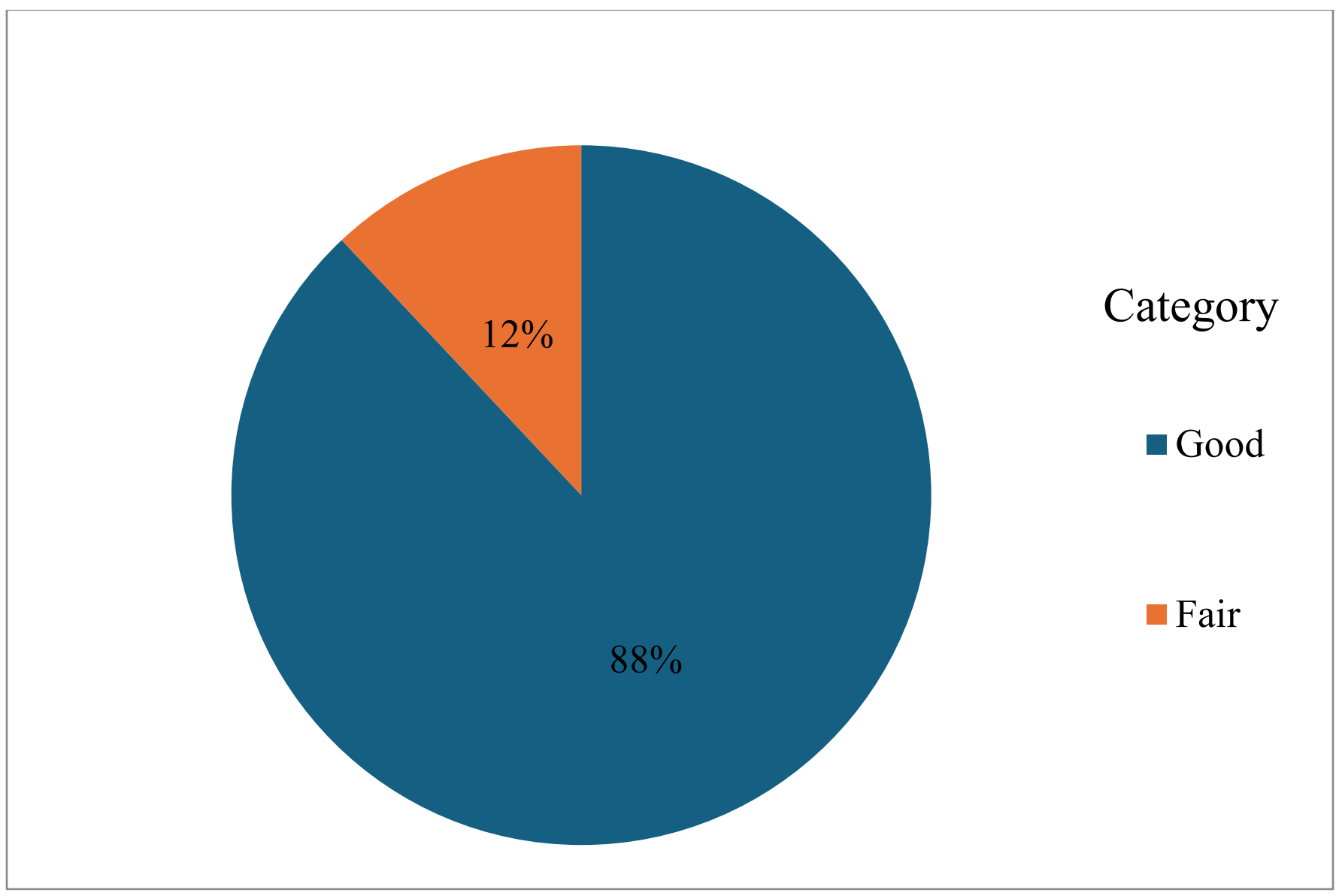

**Fig. 15 QHAdamW prediction per category (PM10 AQI)**

A detailed investigation into the model's sensitivity to learning rates, and weight decay, as well as Epsilon demonstrated that this model has relatively low sensitivity to changes in the learning rate and shows high levels of stable and consistent performance across all settings examined. This study also investigates the ANN model's prediction performance and convergence through MAE, MAPE, MSE, RMSE, NRMSE, WAPE, and R2 values as evaluation metrics. The research showed that the forecasting models for PM2.5 and PM10 have improved reliability/accuracy and convergence data; they can help provide references for understanding how certain model performance impacts (by comparing and evaluating parameters). By providing some important new information about the determinants of forecasting model quality, the study improved our general understanding of air quality forecasting models and facilitated their development.

The performance of Figures 8 through 15 provides a comparative visualization of the forecasting performance of the proposed QHAdamW optimizer in contrast of the original Adam optimizer for predicting AQI values related to PM2.5 and PM10. These figures support the central hypothesis of the study, demonstrating that the QHAdamW optimizer significantly improves convergence, generalization, and predictive accuracy in ANN models trained on noisy urban air quality time series. The trends in forecasting AQI using QHAdamw and Adam are illustrated in Figures 8, 10, 12, and 14. Figures 10 and 14 present the forecasting AQI values for PM2.5 and PM10, respectively, utilizing the QHAdamW optimizer. The predicted values show that QHAdamW effectively captures temporal patterns and the dynamics of pollutants, as they match actual AQI data. The forecasts produced using the original Adam optimizer, on the other hand, are shown in Figures 8 and 12. Figures 9, 11, 13, and 15 show the AQI Category Distribution of Predictions. The model's balanced categorization of the AQI categories results in fewer classifications. Table 16 shows that QHAdamW beat the Adam optimizer with an accuracy rating of 53.05% for PM2.5 and 88.03% for PM10. On the other hand, the category distribution of Adam redictions, as shown in Figures 9 and 13, shows a higher incidence of misclassification and skewed representation of categories, particularly at the baseline AQI values. Figure 8-15 depicts visual evidence that supports the study's quantitative results. The QHAdamW optimizer regularly produces reduced error measurements, such as RMSE and MSE, as well as higher R2 values, suggesting improved model fit and prediction dependability. Furthermore, the enhanced classification of categories strengthens the optimizer's capacity to generalize, which is critical for practical deployment in public health advisories and environmental planning.

## 4. Conclusions and Recommendations

The QHAdamW optimizer is effective in forecasting AQI for PM2.5 and PM10, making a unique addition to air quality modeling in the Philippines. Performane improvements were seen across metrics: for PM2.5, RMSE = 3.38%, MSE = 6.68%, MAE = 2.08%, and predicting accuracy = 53.05%; for PM10, RMSE = 2.70%, MSE = 5.48%, MAE = 1.60%, and accuracy = 88.03%. Training losses decreased to 45.88% and 36.15%, respectively, while convergence also improved. These results suggest that in ANN-based AQI forecasting, QHAdamW is a suitable drop-in replacement for Adam. Consequently, a number of suggestions are put out to expand on these findings. Future research should examine other hyperparameters that might improve model performance, such as batch size, activation functions, and dropout rate. Examining how datasets' features and architectural decisions affect sensitivity and convergence results is also crucial. To evaluate wider application, the QHAdamW optimizer can be expanded to predict additional pollution concentrations, such as sulfur oxides and nitrogen oxides. Additionally, adding model speed to the assessment criteria will guarantee that forecasting systems are both precise and effective for practical implementation. Lastly, proactive decision-making should be supported by the model's predictive in situations involving critical air quality levels.

## References

[1] Frank J. Kelly, and Julia C. Fussell, "Air Pollution and Public Health: Emerging Hazards and Improved Understanding of Risk," *Environmental Geochemistry and Health*, vol. 37, no. 4, pp. 631-649, 2015. [CrossRef] [Google Scholar] [Publisher Link]

[2] Adil Masood, and Kafeel Ahmad, "A Review on Emerging Artificial Intelligence (AI) Techniques for Air Pollution Forecasting: Fundamentals, Application and Performance," *Journal of Cleaner Production*, vol. 322, 2021. [CrossRef] [Google Scholar] [Publisher Link]

[3] Yun Bai et al., "Air Pollutants Concentrations Forecasting using Back Propagation Neural Network based on Wavelet Decomposition with Meteorological Conditions," *Atmospheric Pollution Research*, vol. 7, no. 3, pp. 557-566, 2016. [CrossRef] [Google Scholar] [Publisher Link]

[4] Shirshendu Roy, and Pratyay Mukherjee, "Air Quality Index Forecasting using Hybrid Neural Network Model with Lstm on AQI Sequences," *Proceedings on Engineering Sciences*, vol. 2, no. 4, pp. 431-440, 2020. [CrossRef] [Google Scholar]

[5] Yanlai Zhou, Li-Chiu Chang, and Fi-John Chang, "Explore a Multivariate Bayesian Uncertainty Processor Driven by Artificial Neural Networks for Probabilistic PM2.5 Forecasting," *Science of the Total Environment*, vol. 711, pp. 1-14, 2020. [CrossRef] [Google Scholar] [Publisher Link]

[6] Canyang Guo, Genggeng Liu, and Chi-Hua Chen, "Air Pollution Concentration Forecast Method based on the Deep Ensemble Neural Network," *Wireless Communications and Mobile Computing*, vol. 2020, no. 1, pp. 1-13, 2020. [CrossRef] [Google Scholar] [Publisher Link]

[7] Sales G. Aribe Jr., Bobby D. Gerardo, and Ruji P. Medina, "Time Series Forecasting of HIV/AIDS in the Philippines using Deep Learning: Does COVID-19 Epidemic Matter?," *arXiv preprint*, pp. 144-157, 2022. [CrossRef] [Google Scholar]

[8] Sales G. Aribe Jr., Bobby D. Gerardo, and Ruji P. Medina, "Neural Network-based Time Series Forecasting of HIV Epidemics: The Impact of Antiretroviral Therapies in the Philippines," *Journal of Positive School Psychology*, vol. 6, no. 5, pp. 7844-7855, 2022. [Google Scholar] [Publisher Link]

[9] Alexander Baklanov, and Yang Zhang, "Advances in Air Quality Modeling and Forecasting," *Global Transitions*, vol. 2, pp. 261-270, 2020. [CrossRef] [Google Scholar] [Publisher Link]

[10] Dipayan Guha, Provas Kumar Roy, and Subrata Banerjee, "Application of Backtracking Search Algorithm in Load Frequency Control of Multi-Area Interconnected Power System," *Ain Shams Engineering Journal*, vol. 9, no. 2, pp. 257-276, 2018. [CrossRef] [Google Scholar] [Publisher Link]

[11] Chunhui Li, "RETRACTED: Biodiversity Assessment based on Artificial Intelligence and Neural Network Algorithms," *Microprocessors and Microsystems*, vol. 79, pp. 1-8, 2020. [CrossRef] [Google Scholar] [Publisher Link]

[12] Jolitte A. Villaruz, Bobby D. Gerardo, and Ruji P. Medina, "Philippine Stock Exchange Index Forecasting using a Tuned Artificial Neural Network Model with a Modified Firefly Algorithm," *2023 IEEE 6th International Conference on Pattern Recognition and Artificial Intelligence*, Haikou, China, pp. 1039-1044, 2023. [CrossRef] [Google Scholar] [Publisher Link]

[13] Jolitte A. Villaruz et al., "Scouting Firefly Algorithm and its Performance on Global Optimization Problems," *International Journal of Advanced Computer Science and Applications*, vol. 14, no. 3, pp. 445-451, 2023. [CrossRef] [Google Scholar] [Publisher Link]

[14] Rishika Chauhan et al., "Experimental and Theoretical Evaluation of Thermophysical Properties for Moist Air within Solar Still by using Different Algorithms of Artificial Neural Network," *Journal of Energy Storage*, vol. 30, pp. 1-17, 2020. [CrossRef] [Google Scholar] [Publisher Link]

[15] Jiaxin Yang, and Qiang Long, "A Modification of Adaptive Moment Estimation (Adam) for Machine Learning," *Journal of Industrial and Management Optimization*, vol. 20, no. 7, pp. 2516-2540, 2024. [CrossRef] [Google Scholar] [Publisher Link]

[16] Nuttapong Attrapadung et al., "Adam in Private: Secure and Fast Training of Deep Neural Networks with Adaptive Moment Estimation," *arXiv preprint*, pp. 1-24, 2021. [CrossRef] [Google Scholar] [Publisher Link]

[17] Jerry Ma, and Denis Yarats, "Quasi-Hyperbolic Momentum and Adam for Deep Learning," *arXiv preprint*, pp. 1-38, 2018. [CrossRef] [Google Scholar] [Publisher Link]

[18] Ilya Loshchilov, and Frank Hutter, "Decoupled Weight Decay Regularization," *arXiv preprint*, pp. 1-19, 2017. [CrossRef] [Google Scholar] [Publisher Link]

[19] R. Paje, and J.M. Cuna, "Department of Environment and Natural Resources," *Water Quality Guidelines and General Effluent Standard of 2016*, pp. 1-8, 2016. [Google Scholar]

[20] Environmental Management Bureau - NCR, Regional State of Brown Environment National Capital Region, 2021. [Online]. Available: https://ncr.emb.gov.ph/wp-content/uploads/2021/08/2020-EMB-NCR-RSOBER-COMPRESSED.pdf

[21] Sales G. Aribe Jr., "Explainable Hybrid Framework for Deepfake Detection using Forensic and Deep Neural Features," *SSRN*, pp. 1-19, 2025. [CrossRef] [Google Scholar] [Publisher Link]

[22] Sales Aribe Jr., and Gil Nicholas Cagande, "Benchmarking Federated Learning in Edge Computing Environments: A Systematic Review and Performance Evaluation," *Journal of Advances in Information Technology*, vol. 17, no. 2, pp. 378-389, 2026. [CrossRef] [Google Scholar] [Publisher Link]

[23] Zahraa E. Mohamed, "Using the Artificial Neural Networks for Prediction and Validating Solar Radiation," *Journal of the Egyptian Mathematical Society*, vol. 27, no. 1, pp. 1-13, 2019. [CrossRef] [Google Scholar] [Publisher Link]

[24] Jinghui Chen et al., "Closing the Generalization Gap of Adaptive Gradient Methods in Training Deep Neural Networks," *arXiv preprint*, pp. 1-17, 2018. [CrossRef] [Google Scholar] [Publisher Link]